\documentclass[final]{cvpr}

\usepackage{times}
\usepackage{epsfig}
\usepackage{graphicx}
\usepackage{amsmath}
\usepackage{amssymb}
\usepackage{bbm}
\usepackage{bm}
\usepackage{multirow}
\usepackage[table]{xcolor}
\usepackage[ruled,vlined]{algorithm2e}

\newcommand\blfootnote[1]{%
  \begingroup
  \renewcommand\thefootnote{}\footnote{#1}%
  \addtocounter{footnote}{-1}%
  \endgroup
}

\definecolor{lightgray}{gray}{0.9}

\usepackage[pagebackref=true,breaklinks=true,colorlinks,bookmarks=false]{hyperref}

\def\confYear{CVPR 2021}

\begin{document}

%%%%%%%%% TITLE
\title{Task Programming: Learning Data Efficient Behavior Representations}

\author{Jennifer J. Sun$^\textnormal{1}$ $\quad$
Ann Kennedy$^\textnormal{2}$ $\quad$ 
Eric Zhan$^\textnormal{1}$ $\quad$
David J. Anderson$^\textnormal{1}$ $\quad$
Yisong Yue$^\textnormal{1}$ $\quad$
Pietro Perona$^\textnormal{1}$  
\and
$^\textnormal{1}$Caltech$\quad\quad$ 
$^\textnormal{2}$Northwestern University$\quad\quad$ \\
{\small Code \& Project Website: \url{https://sites.google.com/view/task-programming}}
\vspace{-0.3em}
}

\maketitle

%%%%%%%%% ABSTRACT
%auto-ignore

\begin{abstract}
\vspace{-0.3em}
Specialized domain knowledge is often necessary to accurately annotate training sets for in-depth analysis, but can be burdensome and time-consuming to acquire from domain experts. This issue arises prominently in automated behavior analysis, in which agent movements or actions of interest are detected from video tracking data.
To reduce annotation effort, we present TREBA: a method to learn annotation-sample efficient trajectory embedding for behavior analysis, based on multi-task self-supervised learning.
The tasks in our method can be efficiently engineered by domain experts through a process we call ``task programming'', which uses programs to explicitly encode structured knowledge from domain experts. 
Total domain expert effort can be reduced by exchanging data annotation time for the construction of a small number of programmed tasks. We evaluate this trade-off using data from behavioral neuroscience, in which specialized domain knowledge is used to identify behaviors. We present experimental results in three datasets across two domains: mice and fruit flies.
Using embeddings from TREBA, we reduce annotation burden by up to a factor of 10 without compromising accuracy compared to state-of-the-art features.
Our results thus suggest that task programming and self-supervision can be an effective way to reduce annotation effort for domain experts.
\vspace{-0.3em}
\end{abstract}

%%%%%%%%% BODY TEXT
%auto-ignore

\section{Introduction}

\begin{figure}
    \centering
  \includegraphics[width=\linewidth]{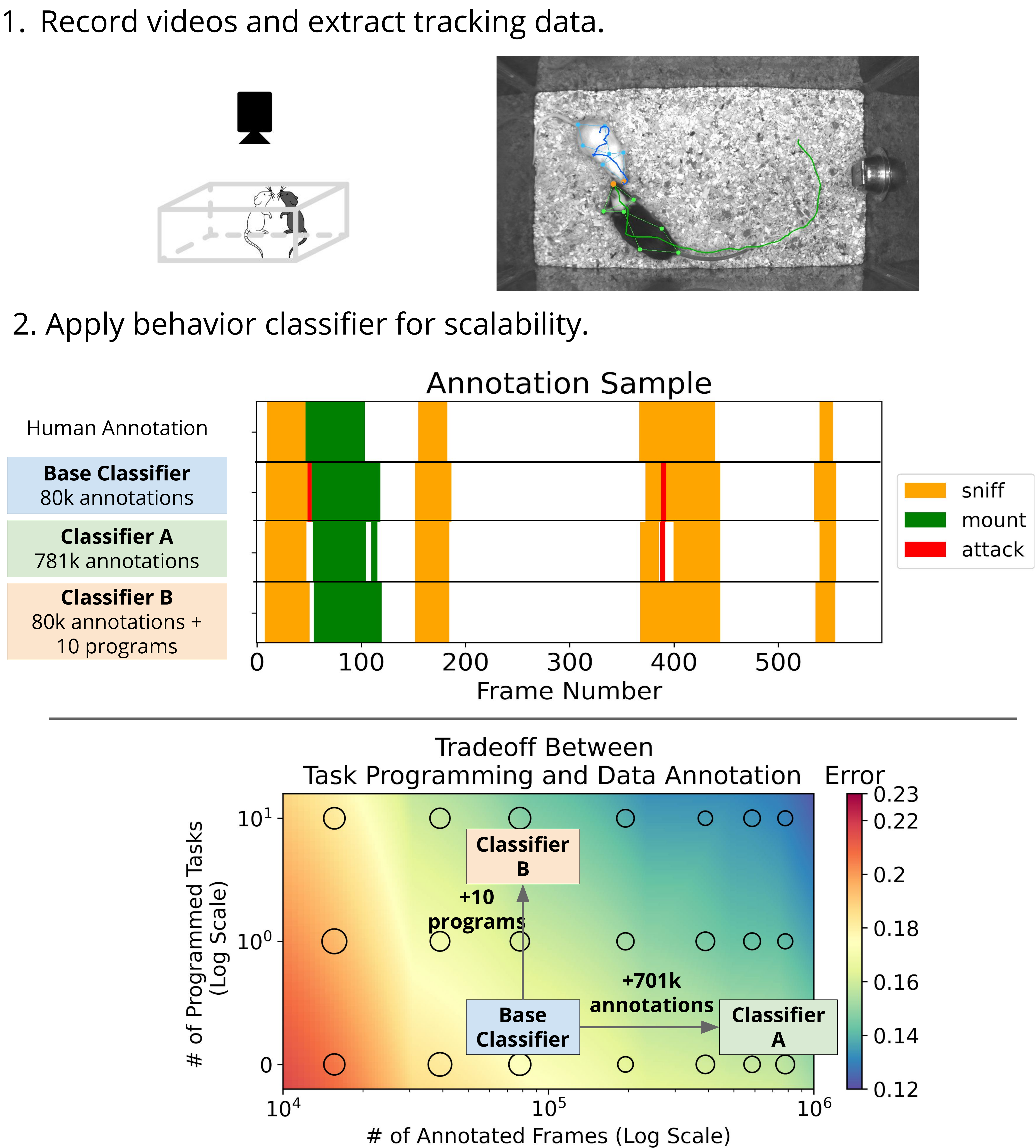}
  \caption{{\bf Overview of our approach.} \textit{Part 1:} A typical behavior study starts with extraction of tracking data from videos. We show 7 keypoints for each mouse, and draw the trajectory of the nose keypoint. \textit{Part 2:} Domain experts can either do data annotation (Classifier A) or task programming (Classifier B) to reduce classifier error. The middle panel shows annotated frames at 30Hz. Colors in the bottom plot represent interpolated performance based on classifier error at the circular markers (full results in Section~\ref{sec:data_efficiency_res}). The size of the marker represents the error variance. }
  \label{fig:introduction}
\end{figure}

Behavioral analysis of one or more agents is a core element in diverse fields of research, including biology~\cite{segalin2020mouse,luxem2020identifying}, autonomous driving~\cite{chang2019argoverse,sun2020scalability}, sports analytics~\cite{yeh2019diverse,zhan2019learning}, and video games~\cite{hofmann2019minecraft,brollcustomizing}.\blfootnote{Correspondence to jjsun@caltech.edu.} In a typical experimental workflow, the location and pose of agents is first extracted from each frame of a behavior video, and then labels for experimenter-defined behaviors of interest are applied on a frame-by-frame basis based on the pose and movements of the agents.
In addition to reducing human effort, automated quantification of behavior can lead to more objective, precise, and scalable measurements compared to manual annotation \cite{anderson2014toward, dell2014automated}. However, training behavior detection models can be data intensive and manual behavior annotation often requires specialized domain knowledge and high-frequency temporal labels. As a result, this process of generating training datasets is time-consuming and effort-intensive for experts. Therefore, methods to reduce annotation effort by domain experts are needed to accelerate behavioral studies.

We study alternative ways for domain experts to improve classifier accuracy beyond simply increasing the sheer volume of annotations.
In particular, we propose a framework that unifies: (1) self-supervised representation learning, and (2) encoding explicit structured knowledge on trajectory data using expert-defined programs. Domain experts can construct these programs efficiently because keypoint trajectories in each frame are typically low dimensional, and experts can already hand-design effective features for trajectory data~\cite{segalin2020mouse,nilsson2020simple}. To best leverage this structured expert knowledge, we develop a framework to learn trajectory representations based on multi-task self-supervised learning, which has not been well-explored for trajectory data. 

\textbf{Our Approach.} Our framework, \textbf{Tr}ajectory \textbf{E}mbedding for \textbf{B}ehavior \textbf{A}nalysis (TREBA), learns trajectory representations through trajectory generation alongside a set of decoder tasks based on expert-engineered programs. These programs are created by domain experts through a process we call task programming, inspired by the data programming paradigm~\cite{ratner2016data}. 
Task programming is a process by which domain experts identify trajectory attributes relevant to the behaviors of interest under study, write programs, and apply those programs to inform representation learning (Section~\ref{sec:methods_tp}). This flexibility in decoder tasks allows our framework to be applicable to a variety of agents and behaviors studied across diverse fields of research.

\textbf{Expert Effort Tradeoffs.} Since task programming will typically require a domain expert’s time, we study the tradeoff between doing task programming and data annotation. We compare behavior classification performance with different amounts of annotated training data and programmed tasks. For example, for the domain illustrated in Figure~\ref{fig:introduction}, domain experts can reduce error by $13\%$ relative to the base classifier by annotating 701k additional frames, or they can reduce error by $16\%$ by learning a representation using 10 programmed tasks in our framework. Our approach allows experts to trade a large number of annotations for a small number of programmed tasks.

We study our approach across two domains in behavioral neuroscience, namely mouse and fly behavior. We chose this setting because it requires specialized domain knowledge for data annotation, and data efficiency is important for domain experts. Furthermore, decoder tasks in our framework can be efficiently programmed by experts based on simple functions describing trajectory attributes for identifying behaviors of interest. 
For example, for mouse social behaviors such as attack~\cite{segalin2020mouse}, important behavior attributes include the speed of each mouse and distance between mice.
The corresponding task could then be to decode these attributes from the learned representations.

Our contributions are:

\begin{itemize}
\item We introduce task programming as an efficient way for domain experts to reduce annotation effort and encode structural knowledge. We develop a novel method to learn an annotation-sample efficient trajectory representation using self-supervision and programmatic supervision. 
\item We study the effect of task programming, data annotation, and different decoder losses on behavior classifier performance.

\item We demonstrate these representations on three datasets in two domains, showing that our method can lead to a $10\times$ annotation reduction for mice, and $2\times$ for flies.
\end{itemize}

%auto-ignore

\section{Related Work}

\textbf{Behavior Modeling}. Behavior modeling using trajectory data is studied across a variety of fields~\cite{luxem2020identifying,chang2019argoverse,sun2020scalability,yeh2019diverse,hofmann2019minecraft,brollcustomizing}. In particular, there is an increasing effort to automatically detect and classify behavior from trajectory data ~\cite{kabra2013jaaba,anderson2014toward, eyjolfsdottir2016learning, mathis2018deeplabcut, egnor2016computational,segalin2020mouse}.
Our experiments are based on behavior classification datasets from behavioral neuroscience~\cite{eyjolfsdottir2014detecting,burgos2012social,segalin2020mouse}, a field where specialized domain knowledge is important for identifying behaviors of interest.

The behavior analysis pipeline generally consists of the following steps: (1) tracking the pose of agents, (2) computing pose-based features, and (3) training behavior classifiers~\cite{burgos2012social,hong2015automated,segalin2020mouse,nilsson2020simple}. To address step 1, there are many existing pose estimation models~\cite{eyjolfsdottir2014detecting,mathis2018deeplabcut,graving2019deepposekit,segalin2020mouse}. In our work, we leverage two existing pose models, \cite{segalin2020mouse} for mice and \cite{eyjolfsdottir2014detecting} for flies, to produce trajectory data. In steps 2 and 3 of the typical behavior analysis pipeline, hand-designed trajectory features are computed from the animals' pose, and classifiers are trained to predict behaviors of interest in a fully supervised fashion~\cite{burgos2012social,hong2015automated,eyjolfsdottir2014detecting,segalin2020mouse}. Training fully supervised behavior classifiers requires time-consuming annotations by domain experts~\cite{anderson2014toward}. Instead, our proposed approach enables domain experts to trade time-consuming annotation work for task programming with representation learning.

\begin{figure*}
    \centering
    \vspace{-0.1in}
  \includegraphics[width=0.85\linewidth]{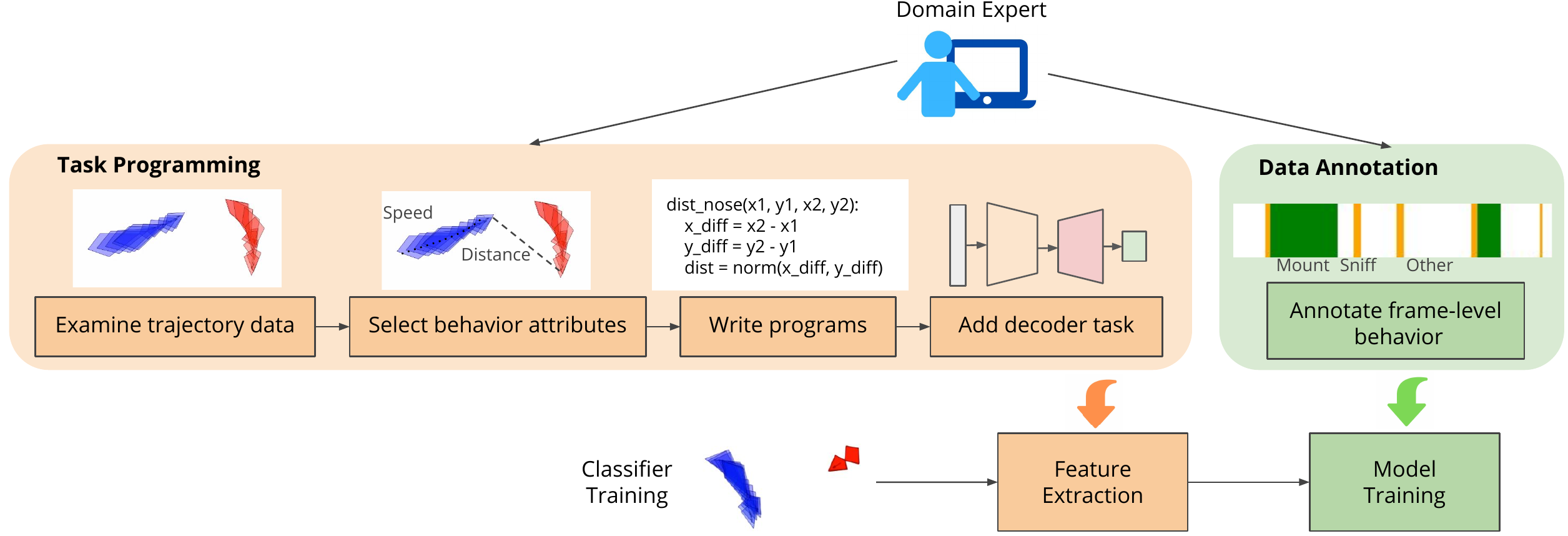}
  \caption{{\bf Task Programming and Data Annotation for Classifier Training.} Domain experts can choose between doing task programming and/or data annotation. Task programming is the process for domain experts to engineer decoder tasks for representation learning. The programs enable learning of annotation-sample efficient trajectory features to improve performance instead of additional annotations.
  \vspace{-0.1in}}
  \label{fig:methods_overall}
\end{figure*}

Another group of work uses unsupervised methods to discover new motifs and behaviors~\cite{hsu2020b,wiltschko2015mapping,berman2014mapping,luxem2020identifying,calhoun2019unsupervised}. Our work focuses on the more common case where domain experts already know what types of actions they would like to study in an experiment.
We aim to improve the data-efficiency of learning expert-defined behaviors.

\textbf{Representation Learning}. Visual representation learning has made great progress in effective representations for images and videos~\cite{goyal2019scaling,gidaris2018unsupervised,chen2020simple,oord2018representation,kolesnikov2019revisiting,han2019video,sun2019videobert}. Self-supervised signals are often used to train this visual representation, such as learning relative positions of image patches~\cite{doersch2015unsupervised}, predicting image rotations~\cite{gidaris2018unsupervised}, predicting future patches~\cite{oord2018representation}, and constrastive learning on augmented images~\cite{chen2020simple}. Compared to visual data, trajectory data is significantly lower dimensional in each frame, and techniques from visual representation learning often cannot be applied directly. For example, while we can create image patches that represent the same visual class, it is difficult to select a partial set of keypoints that represent the same behavior. Our framework builds upon these approaches to learn effective representations for behavioral data.

We investigate different decoder tasks in order to learn an effective behavior representation. One decoder task that we investigate is self-decoding: the reconstruction of input trajectories using generative modeling. Generative modeling has previously been applied to learn representations for visual data~\cite{CycleGAN2017,sun2019videobert,oord2018representation} and language modeling~\cite{radford2018improving}; for trajectory data, we use imitation learning~\cite{wang2017robust,zhan2018generating,zhan2019learning} to train our trajectory representation. 
The other tasks in our multi-task self-supervised learning framework are created by domain experts using task programming (Section~\ref{sec:methods_tp}). This idea of using a human-provided function as part of training has been studied for training set creation \cite{ratner2016data,ratner2017snorkel}, and controllable trajectory generation~\cite{zhan2019learning}. Our work explores these additional decoder tasks to further improve the learned representation over the generative loss alone.

\textbf{Multi-Task Self-Supervised Learning}. We jointly optimize a family of self-supervised tasks in an encoder-decoder setup, making this work an example of multi-task self-supervised learning. Multi-task self-supervised learning has been applied to other domains such as visual data~\cite{doersch2017multi,kolesnikov2019revisiting}, accelerometer recordings~\cite{saeed2019multi}, audio~\cite{ravanelli2020multi} and multi-modal inputs~\cite{shukla2020does,piergiovanni2020evolving}. Generally in each of these domains, tasks are defined ahead of time, as is the case for tasks such as frame reconstruction, colorization, finding relative position of image patches, and video-audio alignment. Most of these tasks are designed for image or video data, and cannot be applied directly to trajectory data. To construct tasks for trajectory representation learning, we propose that domain experts can use task programming to engineer decoder tasks and encode structural knowledge.

%auto-ignore

\section{Methods}

We introduce \textbf{Tr}ajectory \textbf{E}mbedding for \textbf{B}ehavior \textbf{A}nalysis (TREBA), a method to learn an annotation-sample efficient trajectory representation using self-supervision and auxiliary decoder tasks engineered by domain experts. Figure~\ref{fig:methods_overall} provides an overview of the expert's role. In our framework, domain experts replace (a significant amount of) time-consuming manual annotation with the construction of a small number of programmed tasks, reducing total expert effort. Each task places an additional constraint on the learned trajectory embedding. 

\begin{figure*}
    \centering
    %\vspace{-0.1in}
  \includegraphics[width=0.9\linewidth]{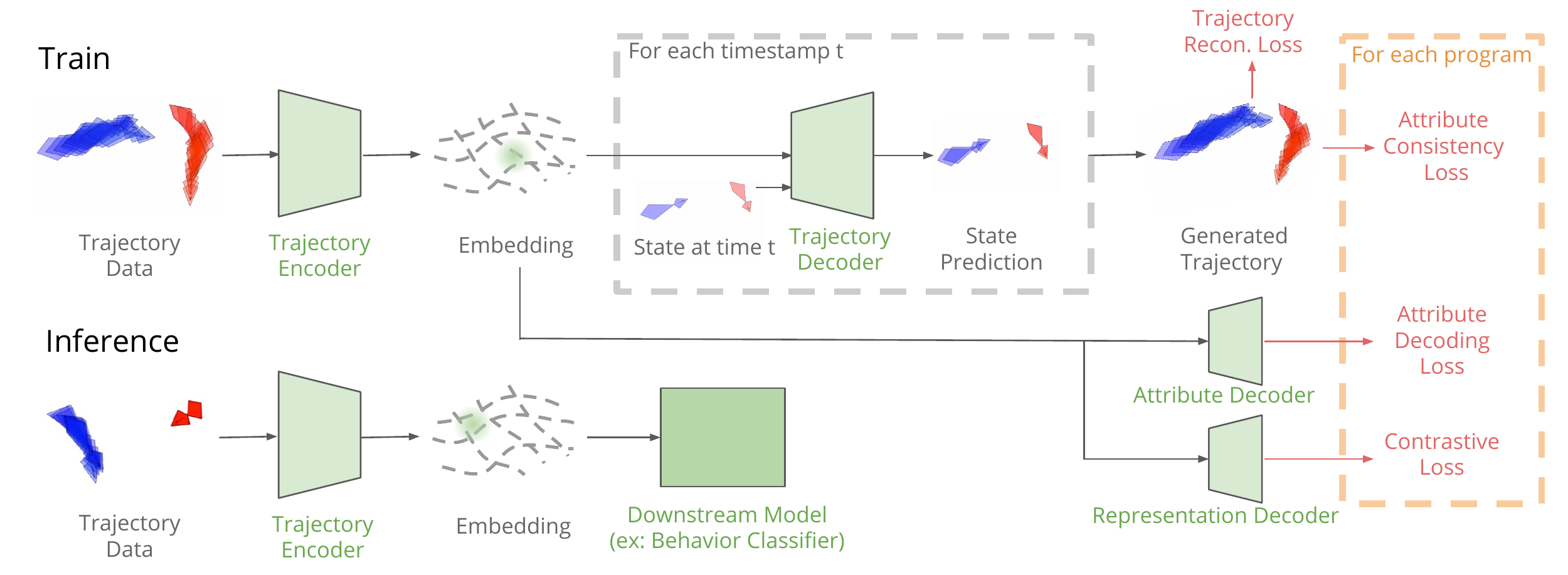}
  \caption{{\bf TREBA Training and Inference Pipelines.} During training, we use trajectory self-decoding and the programmed decoder tasks to train the trajectory encoder. The learned representation is used for downstream tasks such as behavior classification. }
  \label{fig:methods_representation}
\end{figure*}

TREBA uses the expert-programmed tasks based on a multi-task self-supervised learning approach, outlined in Figure~\ref{fig:methods_representation}.
To learn task-relevant low-dimensional representations of pose trajectories, we train a network jointly on (1) reconstruction of the input trajectory (Section~\ref{sec:methods_self_decoding}) and (2) expert-programmed decoder tasks (Section~\ref{sec:methods_mtss}). The learned representation can then be used as input to behavior modeling tasks, such as behavior classification.

\subsection{Trajectory Representations}~\label{sec:methods_self_decoding}
Let $\mathcal{D}$ be a set of $N$ unlabelled trajectories. Each trajectory $\tau$ is a sequence of states $\tau = \{(s_t)\}_{t=1}^T$, where the state $s_i$ at timestep $i$ corresponds to the location or pose of the agents at that timestep. In this study, we divide trajectories from longer recordings into segments of length $T$, but in general trajectory length can vary. For multiple agents, the keypoints of each agent is stacked at each timestep.

Before we introduce our expert-programmed tasks, we will use trajectory reconstruction as an initial self-supervised task. Given a history of agent states, we would like our model to predict the next state. This task is usually studied with sequential generative models. We used trajectory variational autoencoders (TVAEs)~\cite{co2018self,zhan2019learning} to embed the input trajectory using an RNN encoder, $q_{\phi}$, and an RNN decoder, $p_{\theta}$, to predict the next state. The TVAE loss is:
%\vspace{-0.5em}
%
\begin{equation}
\begin{aligned}
\mathcal{L}^{\text{tvae}} = \mathbb{E}_{q_{\phi}} \bigg[ \sum_{t=1}^T -\log(p_{\theta}(s_{t+1} | s_t, \mathbf{z})) \bigg] \\ + D_{KL}(q_{\phi}(\mathbf{z} | \tau ) || p_{\theta}(\mathbf{z})).
\end{aligned}
\end{equation}
We use a prior distribution $p_{\theta}(\mathbf{z})$ on $\mathbf{z}$ to regularize the learned embeddings; in this study, our prior is the unit Gaussian. By optimizing for the TVAE loss only, we learn an unsupervised version of TREBA. When performing subsequent behavior modeling tasks such as classification, we use the embedding mean, $\mathbf{z}_{\mu}$.

\subsection{Task Programming}~\label{sec:methods_tp}
\vspace{-0.005in}
Task programming is the process by which domain experts  create decoder tasks for trajectory self-supervised learning. This process consists of selecting attributes from trajectory data, writing programs, and creating decoder tasks based on the programs (Figure~\ref{fig:methods_overall}). Here, domain experts are people with specialized knowledge for studying behavior, such as neuroscientists or sports analysts.

To start, domain experts identify attributes from trajectory data relevant to the behaviors of interest under study. Behavior attributes capture information that is likely relevant to agent behavior, but is not explicitly included in the trajectory states $\{(s_t)\}_{t=1}^T$. These attributes represent structured knowledge that domain experts are implicitly or explicitly considering for behavior analysis, such as the distance between two agents, agent velocity, or the relative positioning of agent body parts. 

Next, domain experts write a program to compute these attributes on trajectory data, which can be done with existing tools such as MARS~\cite{segalin2020mouse} or SimBA~\cite{nilsson2020simple}. 
Algorithm~\ref{alg:sample_program} shows a sample program from the mouse social behavior domain, for measuring the ``facing angle" between a pair of interacting mice. Each program can be used to construct decoder tasks for self-supervised learning (Section~\ref{sec:methods_mtss}). 

\begin{algorithm}
\SetAlgoLined
 Input: centroid of mouse 1 ($x_1$, $y_1$), centroid of mouse 2 ($x_2$, $y_2$), heading of mouse 1 ($\phi_1$) \\
 $x_{\text{diff}} = x_2 - x_1$ \\
 $y_{\text{diff}} = y_2 - y_1$ \\
 $\theta = \arctan(y_{\text{diff}}, x_{\text{diff}})$ \\
 Return $\theta - \phi_1$
 \caption{Sample Program for Facing Angle}\label{alg:sample_program}
\end{algorithm}

\begin{table}[!t]
  \centering
  \small
  \scalebox{0.9}{
   \begin{tabular}{c | c } 
   \hline
   Domain & Behavior Attributes \\
   \hline
    Mouse & Facing Angle Mouse 1 and 2, Speed Mouse 1 and 2 \\
    & Nose-Nose Distance, Nose-Tail Distance, \\
    & Head-Body Angle Mouse 1 and 2 \\
    & Nose Movement Mouse 1 and 2 \\
    \hline
    Fly & Speed Fly 1 and 2, Fly-Fly Distance \\
    & Angular Speed Fly 1 and 2, Facing Angle Fly 1 and 2 \\
    & Min and Max Wing Angles Fly 1 and 2 \\
    & Major/Minor Axis Ratio Fly 1 and 2 \\
   \hline
\end{tabular}
}
\caption{{\bf Behavior Attributes used in Task Programming.} We base our programmed tasks in our experiments on these behavior attributes from domain experts in each domain.
\vspace{-0.1in}} \label{tab:programmed_tasks}
\end{table}

Our framework is inspired by the data programming paradigm~\cite{ratner2016data}, which applies programs to training set creation. In comparison, our framework uses task programming to unify expert-engineered programs, which encode structured expert knowledge, with representation learning.

Working with domain experts in behavioral neuroscience, we created a set of programs to use in studying our approach. The selected programs are a subset of behavior attributes in \cite{segalin2020mouse} (for mouse datasets) and a subset of behavior attributes in \cite{eyjolfsdottir2014detecting} (for fly datasets). We list the programs used in Table~\ref{tab:programmed_tasks}, and provide more details about the programs in the Supplementary Material.

\subsection{Learning Algorithm}~\label{sec:methods_mtss}
We develop a method to incorporate the programs from domain experts as additional learning signals for TREBA. We consider the following three approaches:
(1) enforcing attribute consistency in generated trajectories (Section~\ref{sec:consistency}), (2) performing attribute decoding directly (Section~\ref{sec:decoding}), (3) applying contrastive loss based on program supervision (Section~\ref{sec:contrastive}). 
Each of these methods applies a different loss on the low-dimensional representation $\mathbf{z}$ of trajectory $\tau$.
Any combinations of these decoding tasks can be combined with self-decoding from Section~\ref{sec:methods_self_decoding} to inform the trajectory embedding $\mathbf{z}$.

\vspace{-0.05in}
\subsubsection{Attribute Consistency}\label{sec:consistency}

Let $\lambda$ be a set of $M$ domain-expert-designed functions measuring agent behavior attributes, such as agent velocity or facing angle. Recall that each $\lambda_j, j = 1...M$ takes as input a trajectory $\tau$, and returns some expert-designed attribute $\lambda_j(\tau)$ computed from that trajectory. For $\lambda_j$ designed for a single frame, we apply the function to the center frame of $\tau$. Attribute consistency aims to maintain the same behavior attribute labels for the generated trajectory as the original. Let $\tilde{\tau}$ be the trajectory generated by the TVAE given the same initial condition as $\tau$ and encoding $\mathbf{z}$.The attribute consistency loss is:
\vspace{-0.05in}
\begin{align}
\mathcal{L}^{\text{attr}} = \mathbb{E}_{\tau \sim \mathcal{D}} \bigg[ \sum_{j=1}^M \mathbbm{1}(\lambda_j(\tilde{\tau})  \neq \lambda_j(\tau)) \bigg].
~\label{eq:consistency}
\end{align}
Here, we show the loss for categorical $\lambda_j$, but in general, $\lambda_j$ can be continuous and any loss measuring differences between $\lambda_j(\tilde{\tau})$ and $\lambda_j(\tau)$ applies, such as mean squared error. We do not require $\lambda$ to always be differentiable, and we use the differentiable approximation introduced in \cite{zhan2019learning} to handle non-differentiable $\lambda$.

\vspace{-0.05in}
\subsubsection{Attribute Decoding}\label{sec:decoding}

Another option is to decode each attribute $\lambda_j(\tau)$ directly from the learned representation $\mathbf{z}$. Here we apply a shallow decoder $f$ to the learned representation, with decoding loss:
\begin{align}
\mathcal{L}^{\text{decode}} = \mathbb{E}_{\tau \sim \mathcal{D}} \bigg[ \sum_{j=1}^M \mathbbm{1}(f(q_{\phi}(\mathbf{z}_{\mu}| \tau))  \neq \lambda_j(\tau)) \bigg].
\end{align}
Similar to Eq. (\ref{eq:consistency}), we show the loss for categorical $\lambda_j$, however any type of $\lambda$ may be used.

\subsubsection{Contrastive Loss}\label{sec:contrastive}

Lastly, the programmed tasks can be used to supervise contrastive learning of our representation. For a trajectory $\tau_i$, and for each $\lambda_j$, positive examples are those trajectories with the same attribute class under  $\lambda_j$. For $\lambda_j$ with continuous outputs, we create a discretized $\hat{\lambda}_j$ in which we apply fixed thresholds to divide the output space into classes. For our work, we apply two thresholds for each program such that our classes are approximately equal in size.

We apply a shallow decoder $g$ to the learned representation, and let $\mathbf{g} = g(q_{\phi}(\mathbf{z}_{\mu}| \tau))$ represent the decoded representation. We then apply the contrastive loss:
\begin{equation}
\begin{aligned}
\mathcal{L}^{\text{cntr.}} = \sum_{i=1}^B \sum_{j=1}^M
\bigg[
\frac{-1}{N_{pos(i,j)}}\sum_{k=1}^{B} \mathbbm{1}_{i\neq k} \cdot \mathbbm{1}_{\lambda_j(\tau_i) = \lambda_j(\tau_k)} \\ 
\cdot \log \frac{\exp(\mathbf{g}_i \cdot  \mathbf{g}_k / t)}{\sum_{l=1}^{N} \mathbbm{1}_{i\neq l} \cdot \exp(\mathbf{g}_i \cdot  \mathbf{g}_l / t)} \bigg],
\end{aligned}
\end{equation}
where $B$ is the batch size, $N_{pos(i,j)}$ is the number of positive matches for $\tau_i$ with $\lambda_j$, and $t>0$ is a scalar temperature parameter. Our form of contrastive loss supervised by task programming is similar to the contrastive loss in ~\cite{khosla2020supervised} supervised by human annotations. A benefit of task programming is that the supervision from programs can be quickly and scalably applied to unlabelled datasets, as compared to expert supervision which can be time-consuming. 
We note that the unsupervised version of this contrastive loss is studied in~\cite{chen2020simple}, based on previous works such as ~\cite{oord2018representation}.

\subsubsection{Data Augmentation}~\label{sec:augmentation}
\vspace{-0.1in}

We can perform data augmentation on trajectory data based on our expert-provided programs. Given the set of all possible augmentations, we define $\Lambda$ to be the subset of augmentations that are \textit{attribute-preserving}: that is, for all $\lambda_j$ in the set of programs, $\lambda_j(\tau) = \lambda_j(\Lambda_m(\tau))$ for some augmentation $\Lambda_m\in\Lambda$. An example of a valid augmentation in the mouse domain is reflection of the trajectory data.

All losses presented above can be extended with data augmentation, by replacing $\tau$ with $\Lambda_m(\tau)$ in losses. For contrastive loss, adding data augmentation corresponds to extending the batch size to $2B$, with $B$ samples from the original and augmented trajectories.

The augmentations we use in our experiments are reflections, rotations, translations, and a small Gaussian noise on the keypoints (mouse data only). In practice, we add the loss for each decoder with and without data augmentation.

%auto-ignore

\section{Experiments}

\subsection{Datasets}

We work with datasets from behavioral neuroscience, where there are large-scale, expert-annotated datasets from scientific experiments. We study behavior for the laboratory mouse and the fruit fly, two of the most common model organisms in behavioral neuroscience.  For each organism, we first train TREBA using large unannotated datasets: for the mouse domain we use an in-house dataset comprised of approximately 100 hours of recorded diadic social interactions (\textbf{Mouse100}), while for the fly domain we use the \textbf{Fly vs. Fly} dataset~\cite{eyjolfsdottir2014detecting} without annotations. 

After pre-training TREBA, we evaluate the suitability of our trajectory representation for supervised behavior classification (classifying frame-level behaviors on continuous trajectory data), on three additional datasets:

\textbf{MARS}. The MARS dataset~\cite{segalin2020mouse} is a recently released mouse social behavior dataset collected in the same conditions as Mouse100.  The dataset is annotated by neurobiologists on a frame-by-frame basis for three behaviors: sniff, attack, and mount. We use the provided train, validation, and test split (781k, 352k, and 184k frames respectively). Trajectories are extracted by the MARS tracker~\cite{segalin2020mouse}.

\textbf{CRIM13}. CRIM13~\cite{burgos2012social} is a second mouse social behavior dataset manually annotated on a frame-by-frame basis by experts. To extract trajectories, we use a version of the the MARS tracker~\cite{segalin2020mouse} fine-tuned on pose annotations on CRIM13. We select a subset of videos from which trajectories can be reliably detected for a train, validation and test split of 407k, 96k, and 142k frames respectively. We evaluated classifier performance on the same three behaviors studied in MARS (sniff, attack, mount).

CRIM13 is a useful test of the robustness of TREBA trained on Mouse100, as the recording conditions in CRIM13 (image resolution $640\times480$, frame rate $25$Hz, and non-centered cage location) are different from those of Mouse100 (image resolution $1024\times570$, frame rate $30$Hz, and centered cage location). 

\textbf{Fly vs. Fly} (Fly). We use the Aggression and Courtship videos from the Fly dataset~\cite{eyjolfsdottir2014detecting}. These videos record interactions between a pair of flies annotated on a frame-by-frame basis for social behaviors by domain experts. Our train, validation and test split has 1067k, 162k, 322k frames respectively. We use the trajectories tracked by \cite{eyjolfsdottir2014detecting} and evaluate on all behaviors with more than 1000 frames of annotations in the full training set (lunge, wing threat, tussle, wing extension, circle, copulation).

\subsection{Training and Evaluation Procedure}

We use the attribute consistency loss (Section~\ref{sec:consistency}) and contrastive loss (Section~\ref{sec:contrastive}) to train TREBA using programs. With the same programs, we find that different loss combinations result in similar performance, and that the combination of consistency and contrastive losses performs the best overall. The results for all loss combinations are provided in the Supplementary Material. 

For the datasets in the mouse domain (MARS and CRIM13) we train TREBA on Mouse100, with 10 programs provided by mouse behavior domain experts. For the Fly dataset, we train TREBA on the training split of Fly without annotations, with 13 programs provided by fly behavior domain experts. The full list is in Table~\ref{tab:programmed_tasks}. We then use the trained encoder, with pre-trained frozen weights, as a trajectory feature extractor over $T=21$ frames, where the representation for each frame is computed using ten frames before and after the current frame. 

We evaluate our classifiers, with and without TREBA features, using Mean Average Precision (MAP). We compute the mean over behaviors of interest with equal weighting. Our classifiers are shallow fully-connected neural networks on the input features. To determine the relationship between classifier performance and training set size, we sub-sample the training data by randomly sampling trajectories (with lengths of 100 frames) to achieve a desired fraction of the training set size. Sampling was performed to achieve a similar class distribution as the full training set. We train each classifier nine times over three different random selections of the training data for each training fraction ($1\%$, $2\%$, $5\%$, $10\%$, $25\%$, $50\%$, $75\%$, $100\%$). Additional implementation details are in the Supplementary Material.

\begin{figure*}

\centering
\includegraphics[width=\textwidth]{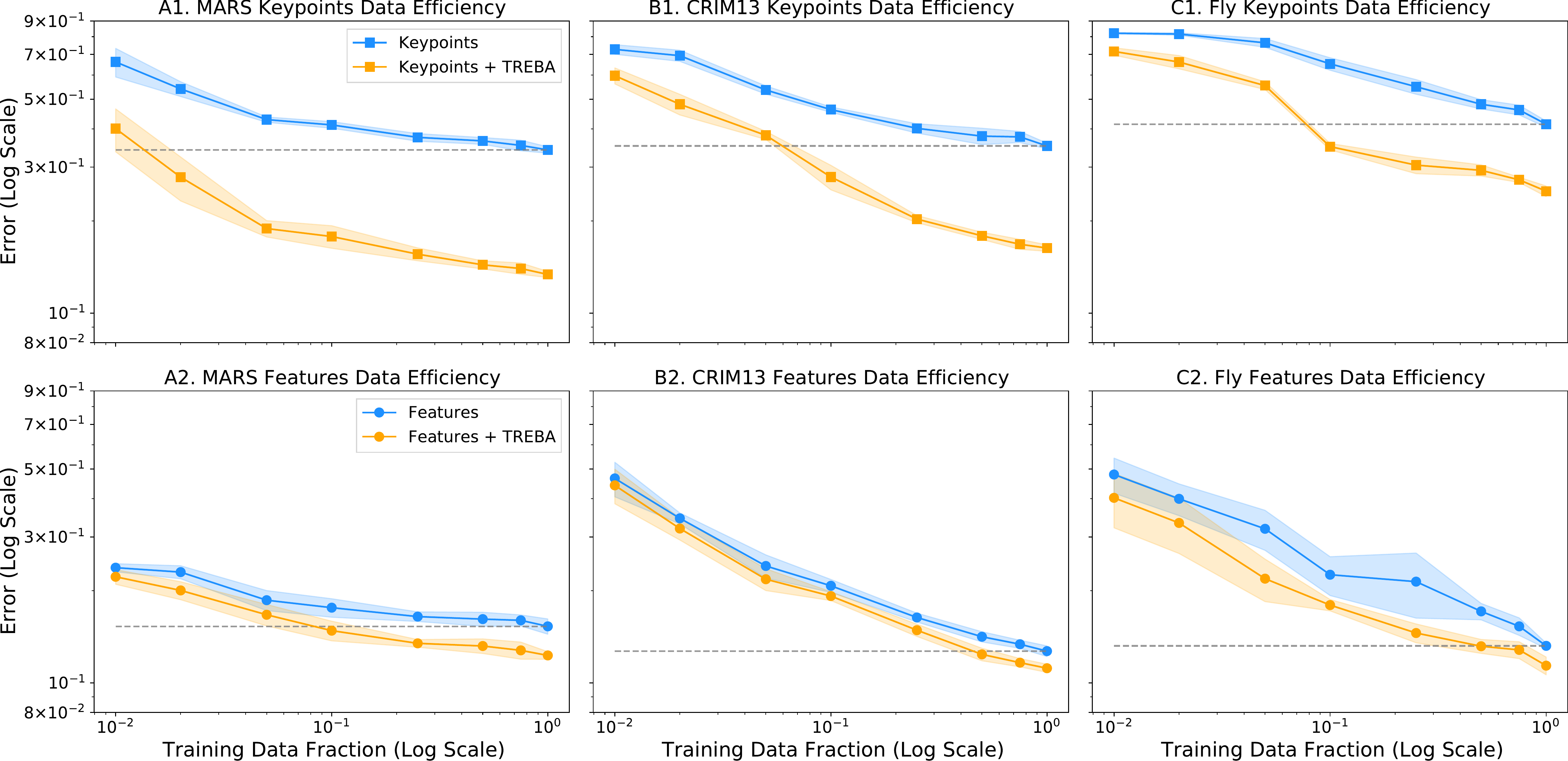}

\caption{ {\bf Data Efficiency for Supervised Classification.} Training data fraction vs. classifier error on MARS (left), CRIM13 (middle) and fly (right). The blue lines represent performance with baseline keypoints and features, and the orange lines are with TREBA. The shaded regions correspond to the classifier standard deviation over nine repeats. The gray dotted line marks the best observed classifier performance when trained on the baseline features (using the full training set). Note the log scale on both the x and y axes.}
\label{fig:efficiency}

\end{figure*}

\subsection{Main Results}~\label{sec:data_efficiency_res}
\vspace{-0.1in}

We evaluate the data efficiency of our representation for supervised behavior classification, by training a classifier to predict behavior labels given both our learned representation and one of either (1) raw keypoints or (2) domain-specific features designed by experts. The TREBA+keypoints evaluation allows us to test the effectiveness of our representation without other hand-designed features, while the TREBA+features evaluation is closer to most potential use cases. The domain-specific features for mice are the trajectory features from~\cite{segalin2020mouse} and features for flies are the trajectory features from~\cite{burgos2012social}. The input features are a superset of the programs we use in Table~\ref{tab:programmed_tasks}.

Our representation is able to improve the data efficiency
for both keypoints and domain-specific features, over all
evaluated amounts of training data availability (Figure~\ref{fig:efficiency}). We discuss each dataset below:

\textbf{MARS}. Our representation significantly improves classification performance over keypoints alone (Figure~\ref{fig:efficiency} A1). We achieve the same performance as the full baseline training using only between $1\%$ and $2\%$ of the data. While this result is partially because our representation contains temporal information, we can also observe a significant increase in data efficiency in A2 compared to domain-specific features, which also contains temporal features. Classifiers using TREBA has the same performance as the full baseline training set with around $5\%\sim10\%$ of data (i.e., $10\times \sim 20\times$ improved annotation efficiency).

\textbf{CRIM13}. We test the transfer learning ability of our representation on CRIM13, a dataset with different image properties than Mouse100, the training set of TREBA. Our representation achieves the same performance as the baseline training with keypoints using around $5\%$ to $10\%$ of the training data (Figure~\ref{fig:efficiency} B1). With domain-specific features, TREBA uses $50\%$ of the data annotation to have the same performance as the full training baseline (i.e., $2\times$ improved annotation efficiency). Our representation is able to generalize to a different dataset of the same organism.

\textbf{Fly}. When using keypoints only, our representation requires $10\%$ of the data (Figure~\ref{fig:efficiency} C1) and for features, our representation requires $50\%$ of the data (Figure~\ref{fig:efficiency} C2) to achieve the same performance as full baseline training. This corresponds to $2\times$ improved annotation efficiency.

\begin{figure*}

\centering
\includegraphics[width=\textwidth]{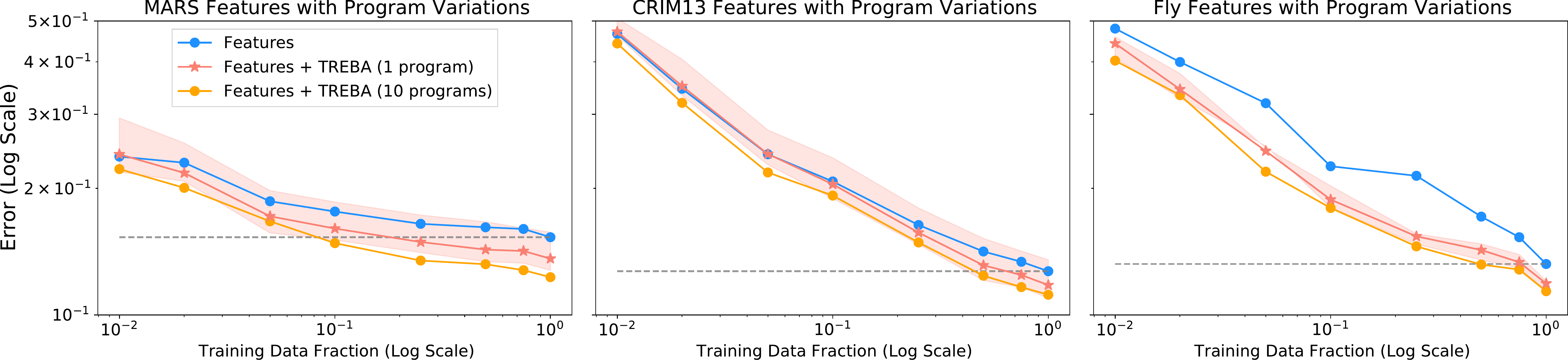}

\caption{ {\bf Varying Programmed Tasks.} Effect of varying number of programmed tasks on classifier data efficiency. The shaded region corresponds to the best and worst classifiers trained using a single programmed task from Table~\ref{tab:programmed_tasks}. The grey dotted line corresponds to the value where the baseline features achieve the best performance (using the full training set). }
\label{fig:program_variations}
\end{figure*}

\subsection{Model Ablations}

We perform the following model ablations to better characterize our approach. In this section, percentage error reduction relative to baseline is averaged over all training fractions. Additional results are in the Supplementary Material.

\textbf{Varying Programmed Tasks}. We test the performance of TREBA trained with each single program provided by the domain experts in Table~\ref{tab:programmed_tasks}, and the average, best, and worst performance is visualized in Figure~\ref{fig:program_variations}. On average, representations learned from a single program is better than using features alone, but using all provided programs further improves performance.

For a single program, there could be a large variation in performance depending on the selected program (Figure~\ref{fig:program_variations}). While the best performing single program is close in classifier MAP to using all programs, the worst performing program may increase error, as in MARS and CRIM13. We further tested the performance using more programs. 

In the mouse domain, we found that with three randomly selected programs, the variation between runs is much smaller compared to single programs (Supplementary Material). With three programs, we achieve comparable average error reduction from baseline features to using all programs (MARS: $14.6\%$ error reduction for 3 programs vs. $15.3\%$ for all, CRIM13: $9.2\%$ for 3 programs vs. $9.5\%$ for all). For the fly domain, we found that we needed seven programs to achieve comparable performance ($20.7\%$ for 7 programs vs. $21.2\%$ for all).

\textbf{Varying Decoder Losses}. When the programmed tasks are fixed, decoder losses with different combinations of consistency (Section~\ref{sec:consistency}), decoding (Section~\ref{sec:decoding}), and contrastive (Section~\ref{sec:contrastive}) loss are similar in performance (Supplementary Material). Additionally, we evaluate the TREBA framework without programmed tasks, with decoder tasks using trajectory generation and unsupervised contrastive loss. While self-supervised representations are also effective at reducing baseline error, we achieve the best classifier performance using TREBA with programmed tasks (Table~\ref{tab:error_reduction_comparison}). Furthermore, we found that training trajectory representations without self-decoding, using the contrastive loss from~\cite{chen2020simple,chen2020big}, resulted in less effective representations for classification (Supplementary Material).

\begin{table}[!t]
  \centering
  \small
  \scalebox{0.93}{
   \begin{tabular}{c | c | c| c} 
   \hline
   & \multicolumn{3}{c}{Keypoint Error Reduction ($\%$)} \\
   Decoder Loss & MARS & CRIM13 & Fly\\
   \hline
    TVAE & $52.2\pm 4.0$ & $34.7\pm 1.5$ & $15.4\pm 2.1$ \\
    \hline
    TVAE+ & \multirow{ 2}{*}{$52.6\pm 3.9$} & \multirow{ 2}{*}{$37.4\pm 2.4$} & \multirow{ 2}{*}{$20.9\pm 1.7$} \\
    Unsup. Contrast & & & \\
   \hline
    TVAE+ & \multirow{ 2}{*}{\bm{$55.1\pm 3.0$}} & \multirow{ 2}{*}{\bm{$41.1\pm 2.1$}} & \multirow{ 2}{*}{\bm{$33.7\pm 1.2$}}\\
    Contrast+Consist & & & \\
    \hline
    \hline
   & \multicolumn{3}{c}{Features Error Reduction ($\%$)} \\
   Decoder Loss & MARS & CRIM13 & Fly\\
   \hline
    TVAE & $13.7 \pm 1.8$ & $8.2\pm 4.6$ & $11.7\pm 4.7$ \\
    \hline
    TVAE+ & \multirow{ 2}{*}{$14.3\pm 2.2$} & \multirow{ 2}{*}{$8.9\pm 4.1$} & \multirow{ 2}{*}{$16.1\pm 1.7$} \\
    Unsup. Contrast & & & \\
   \hline
    TVAE+ & \multirow{ 2}{*}{\bm{$15.3\pm 2.1$}} & \multirow{ 2}{*}{\bm{$9.5\pm 3.8$}} & \multirow{ 2}{*}{\bm{$21.2\pm 4.5$}} \\  
    Contrast+Consist & & & \\     
    \hline
\end{tabular}
}
\caption{{\bf Decoder Error Reductions.} Percentage error reduction relative to baseline keypoints and domain-specific features for training with different decoder losses for TREBA. The average is taken over all evaluated training fractions.
\vspace{-0.1in}} \label{tab:error_reduction_comparison}
\end{table}

\textbf{Data Augmentation}. We removed the losses using the data augmentation described in Section~\ref{sec:augmentation}, and found that performance was slightly lower for all datasets than with augmentation. In particular, adding data augmentation decreases error by $1.2\%$ on MARS, $2.5\%$ on CRIM13, and $5.3\%$ on Fly compared to without data augmentation.

\begin{figure}

\centering
\includegraphics[width=0.85\linewidth]{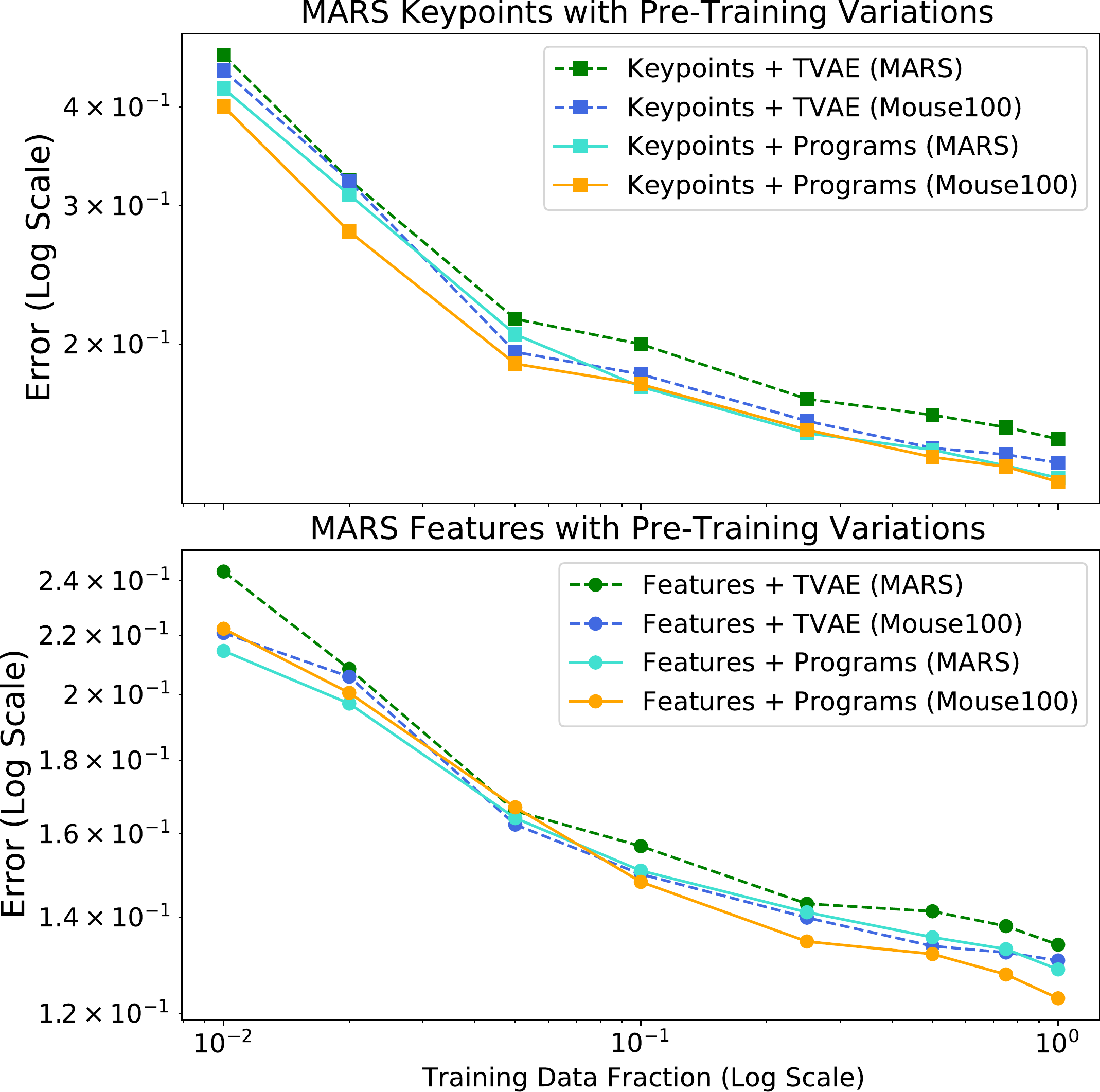}

\caption{ {\bf Pre-Training Data Variations.} Effect of varying pre-training data on classifier data efficiency for the MARS dataset. ``TVAE" corresponds to training TREBA with TVAE losses only, and ``Programs" corresponds to training with all programs.
\vspace{-0.1in}}
\label{fig:pretraining_variations}

\end{figure}

\textbf{Pre-Training Variations} The results shown for MARS was obtained with pre-training TREBA on Mouse100, a large in-house mouse dataset with the same image properties as MARS. Figure~\ref{fig:pretraining_variations} demonstrates the effect of varying TREBA training data amount with TVAE only and with programs. For both keypoints and features, we observe that TVAE (MARS) has the largest error. We see that error can be decreased by either adding more data (features + TVAE (Mouse100) with $3.9\%$ decrease) or adding task programming (features + Programs (MARS) with $4.4\%$ decrease). Adding both more data and task programming results in an average decrease of $5.7\%$ error relative to TVAE (MARS) and the lowest average error.

%auto-ignore

\section{Conclusion}
We introduce a method to learn an annotation-sample efficient Trajectory Embedding for Behavior Analysis (TREBA). To train this representation, we study self-supervised decoder tasks as well as decoder tasks with programmatic supervision, the latter created using task programming.
Our results show that TREBA can reduce annotation requirements by a factor of 10 for mice and 2 for flies. Our experiments on three datasets (two in mice and one in fruit flies) suggest that our approach is effective across different domains. TREBA is not restricted to animal behavior and may be applied to other domains where tracking data is expensive to annotate, such as in sports analytics.

Our experiments highlight, and quantify, the tradeoff between task programming and data annotation. The choice of which is more effective will depend on the cost of annotation and the level of expert understanding in identifying behavior attributes. Directions in creating tools to facilitate program creation and data annotation will help further accelerate behavioral studies.

%auto-ignore

\section{Acknowledgements}

We would like to thank Tomomi Karigo at Caltech for providing the mouse dataset. The Simons Foundation (Global Brain grant 543025 to PP) generously supported this work, and this work is partially supported by NIH Award \#K99MH117264 (to AK), NSF Award \#1918839 (to YY), and NSERC Award \#PGSD3-532647-2019 (to JJS).

{\small
\bibliographystyle{ieee_fullname}
\bibliography{egbib}
}

\clearpage

\onecolumn

\appendix

%auto-ignore
\section*{Supplementary Material}

We provide additional details and experimental results from task programming and TREBA. 
\begin{itemize}
    \item Section~\ref{sec:task_programming_sec} describes the programs we use in the mouse and fly domain (Section~\ref{sec:program_details}) as well as experimental results with varying number of programs (Section~\ref{sec:program_res}).
    \item Section~\ref{sec:implementation_details} provides implementation details on the representation learning architecture for TREBA and behavior classification models.
    \item Section~\ref{sec:additional_res} contains experimental results for decoder loss variations, time estimates, and classification samples.
\end{itemize}

\section{Program Details and Experiments}~\label{sec:task_programming_sec}
\vspace{-0.2in}
\subsection{Program Details}~\label{sec:program_details}
\vspace{-0.2in}

\begin{figure*}

\centering
\includegraphics[width=\textwidth]{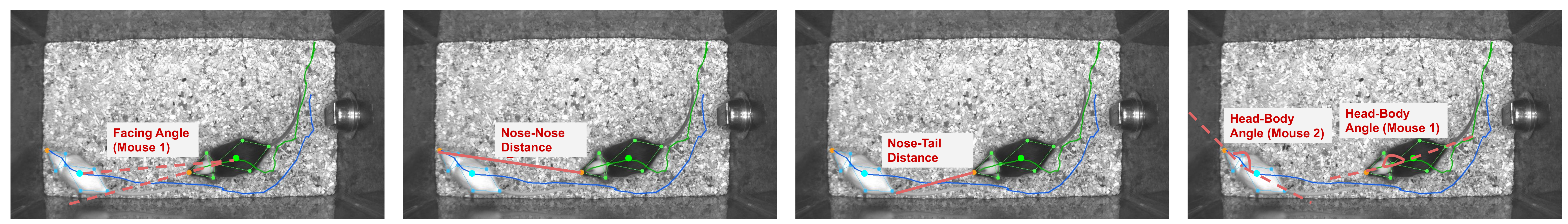}

\caption{Visualizing behavior attributes for mouse dataset.}
\label{fig:mouse_programs}

\end{figure*}

\paragraph{Programs for the Mouse Domain.} We provide additional details on the programs listed in Table 1 in Section 3 of the paper. For the datasets in the mouse domain, programs are selected by domain experts based on the features used for mouse behavior classification in ~\cite{segalin2020mouse}. The experiments are recorded for a standard resident-intruder assay, where an intruder mouse is introduced to the cage of the resident mouse. Mouse 1 corresponds to the resident mouse and mouse 2 corresponds to the intruder mouse. These features are based on the anatomically defined keypoints tracked by the MARS tracker for each mouse: the nose, the ears, the base of the neck, the hips, and the base of the tail. A subset of the programs for the mouse domain is visualized in Figure~\ref{fig:mouse_programs}, and all the programs we use for the mouse domain are listed below.
\begin{itemize}
    \item Facing angle: Relative angle between orientation of the body of the mouse to the line connecting centroids of the two mice. The facing angle is computed for both mice. This describe how closely one mouse is facing the other mouse.
    \item Speed: Change in position of the centroid of the mouse across consecutive frames. The speed is computed for both mice. This property is especially important for helping identify aggressive behavior.
    \item Distance between nose of mouse 1 and 2: Distance between the nose keypoints of each mouse. Distance between noses can be used for identifying when the mice are interacting during social behavior such as sniff. 
    \item Distance between nose of mouse 1 and tail of mouse 2: Distance between the nose keypoint of mouse 1 and base of tail keypoint of mouse 2. Distance between nose of mouse 1 and tail of mouse 2 can be used to identify when the mice are interacting during social behavior such as sniff.
    \item Head-Body Angle: For one mouse, the angle formed by the nose, neck, and base of tail keypoints. This angle is computed for each mouse. This attribute helps describe the body shape of the mouse, since it varies with changes to the relative orientation of the head and body of each mouse.
    \item Nose Movement: Nose movement of each mouse measured by speed relative to the movement of the centroid. This is computed for each mouse. This attribute describes the nose and head speed with respect to the center of the mouse, and can help identify aggressive behavior.
\end{itemize}

\begin{figure*}

\centering
\includegraphics[width=\textwidth]{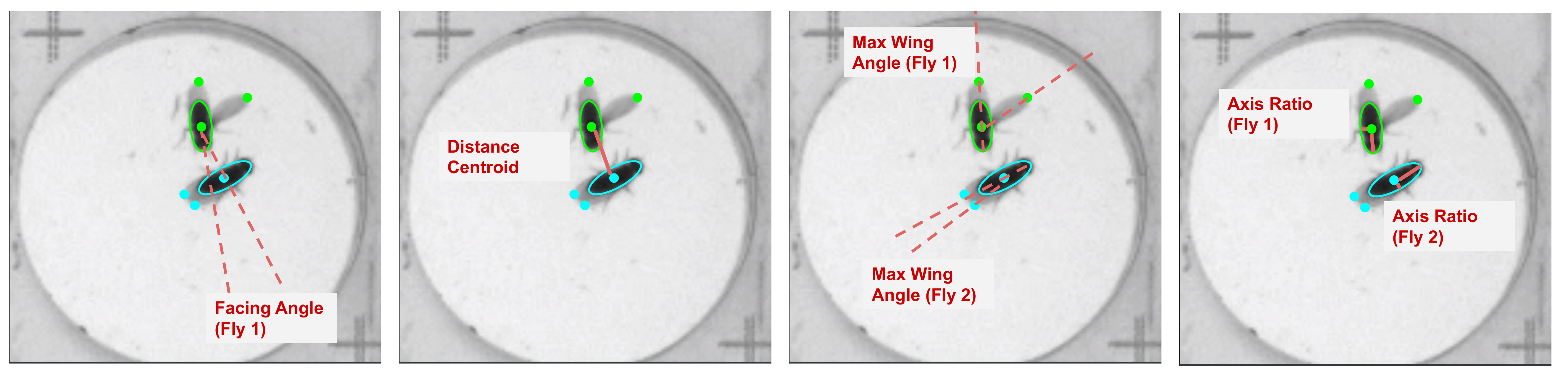}

\caption{Visualizing behavior attributes for fly dataset.}
\label{fig:fly_programs}

\end{figure*}

\paragraph{Programs for the Fly Domain.} For the datasets in the fly domain, these programs are selected by domain experts based on the features used for fly behavior classification in ~\cite{eyjolfsdottir2014detecting}. For each fly, the fly tracker tracks the centroid of the body with a fitted ellipse for the body, the left wingtip keypoint and the right wingtip keypoint. A subset of the programs is visualized in Figure~\ref{fig:fly_programs} and all the programs we use for the fly domain are listed below.
\begin{itemize}
    \item Angular speed: Change in heading direction of the fly across consecutive frames based on the fly body ellipse. The angular speed is computed for all flies. This attribute describes how fast the fly is turning and can help identify behaviors such as tussle and circle.
    \item Minimum and maximum wing angles: The wing angle is the angle between the wing tip keypoint, the centroid, and the point on the back of the body ellipse. Programs are used to compute both the minimum and maximum wing angles for each fly. The wing attributes are especially important behaviors defined by wing position and motion, such as wing extension and wing threat. 
    \item Facing angle: Relative angle between orientation of the body of the fly to the line connecting centroids of the two flies. The facing angle is computed for both flies. The facing angle helps identify if a fly is facing in the direction of the other fly.
    \item Speed: Change in position of the centroid of the fly across consecutive frames. The speed is computed for all flies. This property helps identify behaviors such as lunge, which usually has high speed.
    \item Distance between centroid of fly 1 and 2: The distance between flies is often a good attribute to determine if the flies are interacting during social behavior. 
    \item Ratio between the major and minor axis: The ratio between the major and minor axis length of the fly body ellipse. This is computed for each fly. This attribute is a useful description of the body shape of the fly. When the fly is tilting up, the ratio is usually smaller, and when the fly is flat against the surface, the ratio is usually larger.
\end{itemize}

\subsection{Program Performance Results}~\label{sec:program_res}

We evaluate the representation learned using task programming for each individual program for the mouse and fly domains (Table~\ref{tab:mars_lf_1} to Table~\ref{tab:fly_lf_2}). The evaluation procedure is the same as the main paper, where the MAP is averaged over nine runs with three random selection for each training fraction. MAP@$k\%$ corresponds to the classifier MAP supervised with $k\%$ of the training data. We note that average error reduction discussed in this section is computed across all evaluated training fractions from the main paper ($1\%$, $2\%$, $5\%$, $10\%$, $25\%$, $50\%$, $100\%$).

\begin{table}
\centering
\small
\setlength\tabcolsep{4pt}
\begin{minipage}{0.48\textwidth}
\centering
  \begin{tabular}{l | c c c c c c c c } 
  \hline
  \multicolumn{1}{c|}{MAP@$k\%$} &   $10$  & $50$ & $100$  \\
  \hline
    Domain-specific features & $0.824$ & $0.838$ &  $0.847$ \\
      \rowcolor{lightgray}
  \quad + Head-Body Angle Mouse 1 & $0.853$ & $0.868$ &  $0.874$   \\
  \quad + Facing Angle Mouse 1 &  $0.849$ & $0.866$ & $0.872$   \\
    \rowcolor{lightgray}
  \quad + Distance Nose-Nose & $0.849$ & $0.866$ &  $0.868$  \\   
  \quad + Distance Nose-Tail &  $0.847$ & $0.866$ &  $0.870$  \\     
    \rowcolor{lightgray}
  \quad + Speed Mouse 2 & $0.851$ & $0.866$ &  $0.872$  \\   
  \quad + Head-Body Angle Mouse 2 &  $0.846$ & $0.866$ &  $0.872$  \\   
    \rowcolor{lightgray}
  \quad + Speed Mouse 1 & $0.847$ & $0.862$ & $0.874$  \\
  \quad + Nose Movement Mouse 1 & $0.818$ & $0.841$ &  $0.851$ \\
    \rowcolor{lightgray}
  \quad + Facing Angle Mouse 2 &  $0.814$ & $0.834$ &  $0.843$  \\ 
  \quad + Nose Movement Mouse 2  & $0.820$ & $0.836$ & $0.843$  \\ 
  \hline
  \hline
  \quad + All Programs  & $0.853$ & $0.868$ & $0.877$  \\   \hline
\end{tabular}
\caption{ {\bf Single Program Variations on MARS.} Average MAP of classifiers on MARS trained with features and with TREBA using the specified single program. The order of the programs are based on the average error reduction over all training fractions (highest error reduction at the top). } \label{tab:mars_lf_1}
\end{minipage}%
\hfill
\begin{minipage}{0.48\textwidth}
\centering
  \begin{tabular}{l | c c c c c c c c } 
  \hline
  \multicolumn{1}{c|}{MAP@$k\%$} &   $10$  & $50$ & $100$  \\
  \hline
    Domain-specific features & $0.824$ & $0.838$ &  $0.847$ \\
      \rowcolor{lightgray}
  \quad + 3 programs (A) & $0.856$ & $0.865$ &  $0.872$   \\
  \quad + 3 programs (B) &  $0.850$ & $0.866$ & $0.872$   \\
    \rowcolor{lightgray}
  \quad + 3 programs (C) & $0.855$ & $0.864$ &  $0.878$  \\   
  \hline
\end{tabular}
\caption{ {\bf Additional Program Variations on MARS.} Average MAP of classifiers on MARS trained with features and with TREBA using the three programs. The programs are: (A) Nose Movement Mouse 1, Nose Movement Mouse 2, Facing Angle Mouse 2; (B) Facing Angle Mouse 1, Head-Body Angle Mouse 1, Head-Body Angle Mouse 2; (C) Speed Mouse 1, Nose Movement 1, Distance Nose-Nose. } \label{tab:mars_lf_2}
\end{minipage}
\end{table}

\begin{table}
\centering
\small
\setlength\tabcolsep{4pt}
\begin{minipage}{0.48\textwidth}
\centering
  \begin{tabular}{l | c c c c c c c c } 
  \hline
  \multicolumn{1}{c|}{MAP@$k\%$} &   $10$  & $50$ & $100$  \\
  \hline
    Domain-specific features & $0.792$ & $0.858$ &  $0.873$ \\
      \rowcolor{lightgray}
  \quad + Distance Nose-Nose & $0.811$ & $0.876$ &  $0.889$   \\
  \quad + Distance Nose-Tail &  $0.807$ & $0.881$ & $0.891$   \\
    \rowcolor{lightgray}
  \quad + Speed Mouse 2 & $0.810$ & $0.875$ &  $0.891$  \\   
  \quad + Facing Angle Mouse 1 &  $0.811$ & $0.879$ &  $0.890$  \\     
    \rowcolor{lightgray}
  \quad + Head-Body Angle Mouse 1 & $0.809$ & $0.880$ &  $0.889$  \\   
  \quad + Speed Mouse 1 &  $0.808$ & $0.876$ &  $0.889$  \\   
    \rowcolor{lightgray}
  \quad + Head-Body Angle Mouse 2 & $0.802$ & $0.870$ & $0.881$  \\
  \quad + Nose Movement Mouse 2 & $0.767$ & $0.851$ &  $0.868$ \\
    \rowcolor{lightgray}
  \quad + Nose Movement Mouse 1 &  $0.765$ & $0.852$ &  $0.869$  \\ 
  \quad + Facing Angle Mouse 2  & $0.764$ & $0.848$ & $0.865$  \\ 
  \hline
  \hline
  \quad + All Programs  & $0.808$ & $0.876$ & $0.888$  \\   \hline  
\end{tabular}
\caption{ {\bf Single Program Variations on CRIM13.} Average MAP of classifiers on CRIM13 trained with features and with TREBA using the specified single program. The order of the programs are based on the average error reduction over all training fractions (highest error reduction at the top).} \label{tab:crim13_lf_1}
\end{minipage}%
\hfill
\begin{minipage}{0.48\textwidth}
\centering
  \begin{tabular}{l | c c c c c c c c } 
  \hline
  \multicolumn{1}{c|}{MAP@$k\%$} &   $10$  & $50$ & $100$  \\
  \hline
    Domain-specific features & $0.792$ & $0.858$ &  $0.873$ \\
      \rowcolor{lightgray}
  \quad + 3 programs (A) & $0.811$ & $0.879$ &  $0.889$   \\
  \quad + 3 programs (B) &  $0.810$ & $0.878$ & $0.890$   \\
    \rowcolor{lightgray}
  \quad + 3 programs (C) & $0.807$ & $0.877$ &  $0.887$  \\   
  \hline
\end{tabular}
\caption{ {\bf Additional Program Variations on CRIM13.} Average MAP of classifiers on MARS trained with features and with TREBA using the three programs. The programs are: (A) Nose Movement Mouse 1, Nose Movement Mouse 2, Facing Angle Mouse 2; (B) Facing Angle Mouse 1, Head-Body Angle Mouse 1, Head-Body Angle Mouse 2; (C) Speed Mouse 1, Nose Movement 1, Distance Nose-Nose.} \label{tab:crim13_lf_2}
\end{minipage}
\end{table}

\begin{table}
\centering
\small
\setlength\tabcolsep{4pt}
\begin{minipage}{0.48\textwidth}
\centering
  \begin{tabular}{l | c c c c c c c c } 
  \hline
  \multicolumn{1}{c|}{MAP@$k\%$} &   $10$  & $50$ & $100$  \\
  \hline
    Domain-specific features & $0.774$ & $0.829$ &  $0.868$ \\
      \rowcolor{lightgray}
  \quad + Min. Wing Angle Fly 1 & $0.820$ & $0.864$ &  $0.885$   \\
  \quad + Speed Fly 1 &  $0.818$ & $0.856$ & $0.878$   \\
    \rowcolor{lightgray}
  \quad + Speed Fly 2 & $0.804$ & $0.861$ &  $0.880$  \\   
  \quad + Angular Speed Fly 1 &  $0.821$ & $0.862$ &  $0.881$  \\     
    \rowcolor{lightgray}
  \quad + Max. Wing Angle Fly 2 & $0.814$ & $0.859$ &  $0.882$  \\   
  \quad + Min. Wing Angle Fly 2 &  $0.814$ & $0.859$ &  $0.886$  \\   
    \rowcolor{lightgray}
  \quad + Distance Between Centroids & $0.815$ & $0.858$ & $0.882$  \\
  \quad + Facing Angle Fly 1 & $0.814$ & $0.862$ &  $0.881$ \\
    \rowcolor{lightgray}
  \quad + Axis Ratio Fly 1 &  $0.809$ & $0.855$ &  $0.882$  \\ 
  \quad + Angular Speed Fly 2  & $0.811$ & $0.853$ & $0.879$  \\ 
    \rowcolor{lightgray}
  \quad + Facing Angle Fly 2 &  $0.811$ & $0.853$ &  $0.876$  \\ 
  \quad + Max. Wing Angle Fly 1 & $0.811$ & $0.855$ & $0.883$  \\     \rowcolor{lightgray}
  \quad + Axis Ratio Fly 2 &  $0.797$ & $0.852$ &  $0.880$  \\   
  \hline
  \hline
  \quad + All Programs &  $0.820$ & $0.868$ &  $0.886$  \\   
  \hline
\end{tabular}
\caption{ {\bf Single Program Variations on Fly.} Average MAP of classifiers on Fly trained with features and with TREBA using the specified single program. The order of the programs are based on the average error reduction over all training fractions (highest error reduction at the top).} \label{tab:fly_lf_1}
\end{minipage}%
\hfill
\begin{minipage}{0.48\textwidth}
\centering
  \begin{tabular}{l | c c c c c c c c } 
  \hline
  \multicolumn{1}{c|}{MAP@$k\%$} &   $10$  & $50$ & $100$  \\
  \hline
    Domain-specific features & $0.774$ & $0.829$ &  $0.868$ \\
      \rowcolor{lightgray}
  \quad + 3 programs (A) & $0.814$ & $0.857$ &  $0.880$   \\
  \quad + 3 programs (B) &  $0.814$ & $0.857$ & $0.878$   \\
    \rowcolor{lightgray}
  \quad + 3 programs (C) & $0.815$ & $0.863$ &  $0.880$  \\     \quad + 7 programs (A) & $0.819$ & $0.869$ &  $0.889$   \\
      \rowcolor{lightgray}
  \quad + 7 programs (B) &  $0.820$ & $0.863$ & $0.885$   \\
  \quad + 7 programs (C) & $0.815$ & $0.860$ &  $0.882$  \\   
  
  \hline
\end{tabular}
\caption{ {\bf Additional Program Variations on Fly}. Average MAP of classifiers on Fly trained with features and with TREBA using three and seven programs. The sets of three programs are: (A) Speed Fly 1, Min/Max Wing Angle Fly 1; (B) Speed Fly 2, Facing Angle Fly 1, Axis Ratio Fly 2; (C) Min Wing Angle Fly 2, Speed Fly 1, Axis Ratio Fly 1. \\
The sets of seven programs are: (A) Distance, Angular Speed Fly 1/2, Max Wing Angle Fly 1, Min/Max Wing Angle Fly 2, Facing Angle Fly 1; (B) Distance, Speed/Angular Speed Fly 1, Max Wing Angle Fly 1, Facing Angle Fly 2, Axis Ratio Fly 1/2; (C) Speed Fly 2, Min/Max Wing Angle Fly 1/2, Facing Angle Fly 1/2.} \label{tab:fly_lf_2}
\end{minipage}
\end{table}

\paragraph{Mouse Program Evaluations.} We train TREBA with each individual program from the mouse domain on Mouse100 using the consistency and contrastive losses, and evaluate the performance of domain-specific features+TREBA on MARS (Table~\ref{tab:mars_lf_1}) and CRIM13 (Table~\ref{tab:crim13_lf_1}). In general, TREBA trained on a single program improves classifier performance, except for the bottom three programs and there is a high variation in performance (for example, Nose Movement Mouse 1 vs. Distance Nose-Nose). The two distance attributes, Head-Body Angle of Mouse 1, and Facing Angle of Mouse 1 generally performs the best across MARS and CRIM13 as a single program.  

The best performing single program, Head-Body Angle Mouse 1 for MARS (Table~\ref{tab:mars_lf_1}) and Distance Nose-Nose for CRIM13 (Table~\ref{tab:crim13_lf_1}), is comparable in performance to using all programs. When comparing average error reduction from baseline hand-designed features on MARS, TREBA using the top single program (Head-Body Angle Mouse 1, Table~\ref{tab:mars_lf_1}) achieves an error reduction of $14.5\%$ and using all programs achieves an error reduction of $15.3\%$. On CRIM13, the top single program (Distance Nose-Nose, Table~\ref{tab:crim13_lf_1}) achieves an error reduction of $9.3\%$ and all programs achieves an error reduction of $9.5\%$. In contrast, the worst performing single program may reduce performance in the mouse domain. We study whether this performance variance can be reduced by adding more programs.

We experiment training TREBA using sets of three programs and evaluating behavior classification performance (MARS in Table~\ref{tab:mars_lf_2}, CRIM13 in Table~\ref{tab:crim13_lf_2}). We note that program sets B and C are three randomly selected programs from the full mouse program list, and program set A consists of the worst performing three single programs. Despite program set A consisting of the lowest performing single programs, we see that for both MARS and CRIM13, the different sets of three programs are similar in performance and is comparable to using all programs (Table~\ref{tab:mars_lf_2}, Table~\ref{tab:crim13_lf_2}). By training TREBA with three programs instead of one, the performance variation across program sets is much lower. We recommend that domain experts train with multiple programs, unless the best performing single program is known.

\paragraph{Fly Program Evaluations.} We train TREBA on the Fly dataset without annotations using individual programs from the fly domain and evaluate on the Fly dataset (Table~\ref{tab:fly_lf_1}). Training with TREBA with any single expert-engineered program improves performance in the fly domain. We see that the speed and wing angle features generally perform the best. The top performing single program (Min. Wing Angle Fly 1, Table~\ref{tab:fly_lf_1}) achieves $19.4\%$ average error reduction over baseline features, comparing to $21.2\%$ using all programs. Similar to the mouse domain, if the best performing program is known ahead of time, we can achieve comparable performance to training TREBA using all programs. However, the best and worst single programs have a large variance, and we experiment with adding more programs.

We start by training on sets of three programs, and found that there is a gap in performance to using all programs (Table~\ref{tab:fly_lf_2}). We additionally experiment with sets of seven programs to close this performance gap. Training with randomly selected three or seven programs (Table~\ref{tab:fly_lf_2}) has much smaller variations across program selections compared to single programs (Table~\ref{tab:fly_lf_1}). For the fly domain, we found that training with seven programs is able to achieve comparable performance to using all programs.

\section{Additional Implementation Details}~\label{sec:implementation_details}

We provide hyperparameters used in training TREBA (Table~\ref{tab:hyperparameter_1}) and the classification models (Table~\ref{tab:hyperparameter_2}). Our code is available at \url{https://github.com/neuroethology/TREBA}.

For training TREBA, the TVAE consists of a bi-directional GRU with 256 units for the encoder, followed by linear layers, with a latent dimension of 32. We take the encoding mean from the encoder, $\mathbf{z}_{\mu}$, as our learned representation of the trajectory. The decoder for the self-decoding task is also a GRU with 256 units followed by linear layers to predict the state in the next timestamp (by predicting the change from the current state to the next state). The decoder for the other tasks (attribute decoding, contrastive loss) consists of a fully connected neural network with 32 units. For the attribute consistency loss, we use the method proposed in \cite{zhan2019learning} to train a 256 unit GRU to approximate non-differentiable programs. When the corresponding decoder is used, we weigh the consistency loss by 1.0, contrastive loss by 10.0, and the decoding loss by 1.0. We train TREBA using Adam optimizer with a learning rate of 0.0002. The input trajectories to the TREBA model are the detected keypoints for mouse and fly using the MARS tracker~\cite{segalin2020mouse} and Fly tracker~\cite{eyjolfsdottir2014detecting} respectively. At each frame, we stack the keypoints of the agents, and a trajectory in our experiment consists of 21 frames. We normalize the coordinates of pose keypoints by the image pixel dimensions.

For classification, we use a shallow fully connected network with two hidden layers. We decrease the size of the network as the input training fraction decreases (Table~\ref{tab:hyperparameter_2}). The model size and other hyperparameters are chosen based on the validation split. Our results are all reported on the test split.
We train the classification models using cross-entropy loss and Adam optimizer with learning rate 0.001.

\begin{table}[!t]
  \centering
  \small
  \scalebox{1.0}{
   \begin{tabular}{c | c | c| c | c| c | c } 
   \hline
   Dataset & Batch size & $\textbf{z}$-dim & Encoder Units & Decoder Units & Temperature $t$ & Learning Rate \\
   \hline
    Mouse100 & 128 &  32 & 256 & 256 & 0.07 & 0.0002 \\
    \hline
    Fly & 128 & 32 & 256 & 256  & 0.07 & 0.0002\\
   \hline
\end{tabular}
}
\caption{{\bf Hyperparameters for Representation Learning.}} \label{tab:hyperparameter_1}
\end{table}

\begin{table}[!t]
  \centering
  \small
  \scalebox{1.0}{
   \begin{tabular}{c | c | c| c | c| c } 
   \hline
   Dataset & Batch size & Classifier Units  & Classifier Units  & Classifier Units & Learning Rate  \\
    & & ($100\%$, $75\%$, $50\%$) & ($25\%$, $10\%$) & ($5\%$, $2\%$, $1\%$) & \\
   \hline
    MARS &  512 & 256, 32 & 128, 16 & 64, 16 & 0.001 \\
    \hline
    CRIM13 &  512 & 256, 32 & 128, 16 & 64, 16 & 0.001  \\
    \hline
    Fly &  512 & 256, 32 & 128, 16 & 64, 16 & 0.001  \\
   \hline
\end{tabular}
}
\caption{{\bf Hyperparameters for Classification Models.}} \label{tab:hyperparameter_2}
\end{table}

\section{Additional Experimental Results}~\label{sec:additional_res}

\subsection{Decoder Loss Variations}~\label{sec:decoder_loss}

We evaluate TREBA trained using different decoder losses on supervised behavior classification. The procedure is the same as described in the main paper. We evaluate performance given both our learned representation and one of either (1) raw keypoints or (2) domain-specific features designed by experts. The input keypoints to the classification model are the detected poses from the MARS tracker~\cite{segalin2020mouse} and the Fly tracker~\cite{eyjolfsdottir2014detecting}. The input domain-specific features are the hand-designed trajectory features for mouse~\cite{segalin2020mouse} and fly~\cite{eyjolfsdottir2014detecting}. The input features are a superset of the programs we use to train TREBA (listed in Table 1 from the main paper and described in Supplementary Material Section~\ref{sec:program_details}).

We compare the MAP of TREBA representations trained with different decoder losses in Table~\ref{tab:efficiency_losses}. The rows for TVAE and TVAE + Unsup. Contrast represents TREBA trained without programmed tasks and the remaining rows represents different combinations of decoder losses with programmed tasks. The average error reduction of these runs from baseline are shown in Table 2 in the main paper. Across all domains and training data amounts, we see that the learned representation improves classifier performance for both keypoints and domain-specific features. The improvements in performance are generally larger when we use keypoints, most likely because domain-specific features already contain informative features for classification.  Furthermore, we experiment training TREBA without self-decoding, using unsupervised contrastive loss similar to~\cite{chen2020simple,chen2020big}, and we see that the classifier MAP is lower than training with self-decoding using the TVAE loss.

%We found that using different projection layer depths as in~\cite{chen2020simple,chen2020big} performed similarly.
%, and here we show the results from using a projection layer with three fully connected layers

Table~\ref{tab:efficiency_losses} demonstrates that when using task programming, different decoder loss combinations (attribute consistency, decoding, and contrastive loss) are similar in performance in general, except when consistency loss is used alone. The lowest performing loss is when we use attribute consistency loss alone, which is applied to the generated trajectory and not directly to the representation. This result suggests that having at least one loss term directly applied on the representation (either decoding or contrastive loss) is beneficial.

Comparing task programming with TVAE loss alone, we see that TVAE loss is generally lower in performance (Table~\ref{tab:efficiency_losses}, Table 2 in the main paper). We note that TVAE loss alone corresponds to self-supervised learning with self-decoding only. Adding unsupervised contrastive loss to self-decoding improves performance relative to the TVAE, but we can improve the performance further using losses based on task programming (for example, TVAE+Contrastive+Consistency). 

\begin{table*}[!t]
  \centering
  \small
  \scalebox{1.0}{
  \begin{tabular}{l | c c c | c c c | c c c} 
  \hline
  \multicolumn{1}{c|}{Dataset} & \multicolumn{3}{c|}{MARS} & \multicolumn{3}{c|}{CRIM13} & \multicolumn{3}{c}{Fly} \\
  \multicolumn{1}{c|}{MAP@$k\%$} & $10$ & $50$ & $100$ &  $10$ & $50$ & $100$ & $10$ & $50$ & $100$ \\
  \hline
    Keypoints &  $0.588$ & $0.635$ &  $0.656$ &  $0.538$ & $0.621$ & $0.648$ & $0.348$ & $0.519$ & $0.586$\\
      \rowcolor{lightgray}
  \quad + TVAE &  $0.817$ & $0.852$ &  $0.859$ &  $0.703$ & $0.796$ & $0.820$ & $0.419$ & $0.635$ & $0.722$  \\
  \quad + TVAE+Unsup. Contrast &  $0.815$ & $0.852$ &  $0.866$ &  $0.706$ & $0.813$ & $0.837$ & $0.521$ & $0.667$ & $0.739$  \\
  \rowcolor{lightgray}
  \quad + TVAE+Consist  &  $0.704$ & $0.763$ & $0.776$  &  $0.581$ & $0.694$ & $0.720$ & $0.497$ & $0.657$ & $0.729$ \\
  \quad + TVAE+Contrast &  $0.804$ & $0.851$ &  $0.868$ & $0.707$ & $0.813$ & $0.838$ &  $0.625$ & $0.712$ & $0.753$ \\  
  \rowcolor{lightgray}
  \quad + TVAE+Decode & $0.825$ & $0.857$ &  $0.872$ & $0.719$  & $0.828$ & $0.848$ & $0.666$ & $0.737$ & $0.773$ \\   
  \quad + TVAE+Contrast+Consist & $0.822$ & $0.856$ &  $0.866$ &  $0.722$ & $0.821$ & $0.837$ &  $0.650$ & $0.707$ & $0.750$ \\
  \rowcolor{lightgray}
  \quad + TVAE+Decode+Consist & $0.820$ & $0.855$ &  $0.870$ &  $0.707$ & $0.813$ & $0.837$ & $0.432$ & $0.603$ & $0.688$ \\
  \quad + TVAE+Contrast+Decode &  $0.821$ & $0.859$ &  $0.871$ & $0.693$ & $0.811$  &  $0.830$ & $0.645$ & $0.738$ & $0.775$ \\  
  \rowcolor{lightgray}
  \quad + TVAE+Contrast+Decode+Consist & $0.821$ & $0.857$ & $0.870$ & $0.693$ & $0.811$ & $0.834$ & $0.484$ & $0.616$ & $0.679$\\ 
   \quad + Unsup. Contrast & $0.582$ & $0.704$ & $0.735$ & $0.587$ & $0.697$ & $0.721$ & $0.384$ & $0.560$ & $0.645$ \\  
  \hline 
  \hline
    Domain-specific features & $0.824$ & $0.838$ &  $0.847$ &  $0.792$ & $0.858$ & $0.873$ & $0.774$ & $0.829$ & $0.868$\\
      \rowcolor{lightgray}
  \quad + TVAE & $0.850$ & $0.866$ &  $0.869$  & $0.808$ & $0.874$ & $0.885$ & $0.791$ & $0.852$ & $0.880$ \\
  \quad + TVAE+Unsup. Contrast & $0.850$ & $0.866$ &  $0.871$  & $0.808$ & $0.876$ & $0.889$ & $0.811$ & $0.858$ & $0.882$ \\ 
  \rowcolor{lightgray}
  \quad + TVAE+Consist &  $0.824$ & $0.841$ & $0.853$  &  $0.775$ & $0.854$ & $0.871$ & $0.812$ & $0.856$ & $0.882$ \\
  \quad + TVAE+Contrast & $0.853$ & $0.868$ &  $0.872$ & $0.811$ & $0.876$ & $0.889$ &   $0.834$ & $0.869$ & $0.888$  \\   
  \rowcolor{lightgray}
  \quad + TVAE+Decode &  $0.851$ & $0.869$ &  $0.874$  & $0.805$ & $0.880$ & $0.892$ &  $0.815$ & $0.866$ & $0.883$ \\   
  \quad + TVAE+Contrast+Consist & $0.853$ & $0.868$ &  $0.877$ &  $0.808$ & $0.876$ & $0.888$ &  $0.820$ & $0.868$ & $0.886$ \\
  \rowcolor{lightgray}
  \quad + TVAE+Decode+Consist & $0.848$ & $0.868$ &  $0.873$ &  $0.808$ & $0.872$ & $0.888$ & $0.781$ & $0.838$ & $0.862$ \\   
  \quad + TVAE+Contrast+Decode &  $0.846$ & $0.867$ &  $0.876$ & $0.811$ & $0.878$  &  $0.892$ & $0.810$ & $0.862$ & $0.885$  \\ 
  \rowcolor{lightgray}
  \quad + TVAE+Contrast+Decode+Consist  & $0.851$ & $0.866$ & $0.872$ & $0.813$ & $0.879$ & $0.892$ & $0.783$ & $0.842$ & $0.868$ \\ 
  \quad  + Unsup. Contrast & $0.830$ & $0.847$ & $0.853$ & $0.787$ & $0.858$ & $0.874$ & $0.784$ & $0.842$ & $0.866$\\  
  \hline
\end{tabular}
}
\caption{\textbf{Decoder Loss Variations.} Comparing data efficiency of TREBA trained with different decoder losses with respect to classifier MAP. Keypoints and domain-specific features represent baseline input features. TVAE represents training TREBA with self-decoding only and contrastive, decoding and consistency loss are described in Section 3.3 of the main paper.} \label{tab:efficiency_losses}
\end{table*}

\paragraph{Random Program Inputs.} We further experiment with training TREBA (Contrastive + Consistency), without expert-engineered programs, using a program that returns one of three classes randomly with equal probability for each trajectory. We found that the error reduction when using a program with random outputs is between training using TVAE alone and using unsupervised contrastive loss. On MARS relative to baseline features, TREBA with a random program achieves an error reduction of $14.0\%$, compared to $13.7\%$ for TVAE, $14.3\%$ for TVAE + Unsup. Contrastive, $15.3\%$ for all programs (Table 2 in main paper). On Fly relative to baseline features, TREBA with a random program achieves an error reduction of $13.6\%$, compared to $11.7\%$ for TVAE, $16.1\%$ for TVAE + Unsup. Contrastive, $21.2\%$ for all programs (Table 2 in main paper). We tried adding more random programs during training, but did not observe an increase in performance. The lower performance of TREBA using random program inputs compared to all programs suggests that programs engineered using structured expert knowledge based on behavior attributes is important for improving the effectiveness of the learned representation.

\subsection{Time Estimates}~\label{sec:time_estimates} 

Based on domain expert estimates in neurobiology, behavior annotation takes 4 times the length of 30Hz videos, while task programming takes 5 to 10 minutes per program. We note that this time estimate is from domain experts familiar with data annotation and trajectory feature design. Applying this estimate to Figure 4 A2 B2 C2 in the main paper (for data efficiency with domain-specific features), in regions of low time investment ($<2$ hours), it is generally better for domain experts to annotate more data. For performance at $>2$ hours, task programming provides a better return on investment of the expert's time. Task programming requires an initial effort to produce the programs, which then scales to any data amount with no additional effort. Note that this time is variable depending on the domain expert and the domain, and our estimate is based on neurobiologists familiar with data annotation and trajectory feature design.

\subsection{Classification Samples}~\label{sec:samples}

\begin{figure*}

\centering
\includegraphics[width=\textwidth]{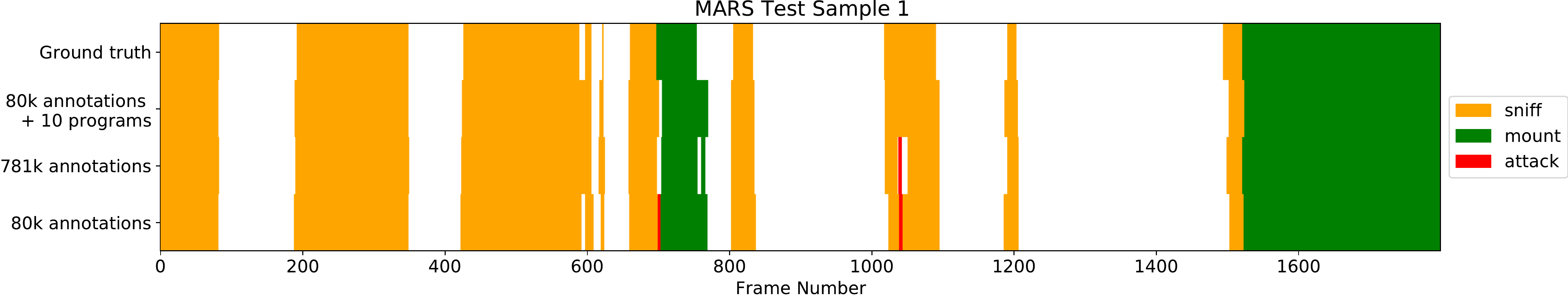}

\includegraphics[width=\textwidth]{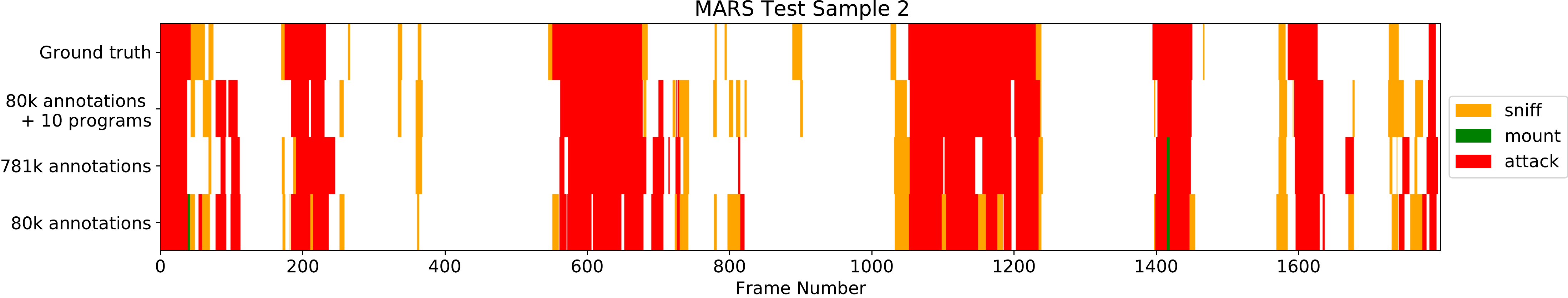}

\caption{ \textbf{Annotations on MARS Test Set.} Classifier annotations on MARS dataset vs. ground truth at 30Hz. For each sample, the second row corresponds to features + TREBA trained with 10 expert-engineered programs, with $\sim 10\%$ of the supervised behavior annotations. The third row corresponds to training using domain-specific features on the full dataset. The last row corresponds to training using domain-specific features on $\sim 10\%$ of data, without TREBA.}
\label{fig:sample_1}

\end{figure*}

\begin{figure*}

\centering
\includegraphics[width=\textwidth]{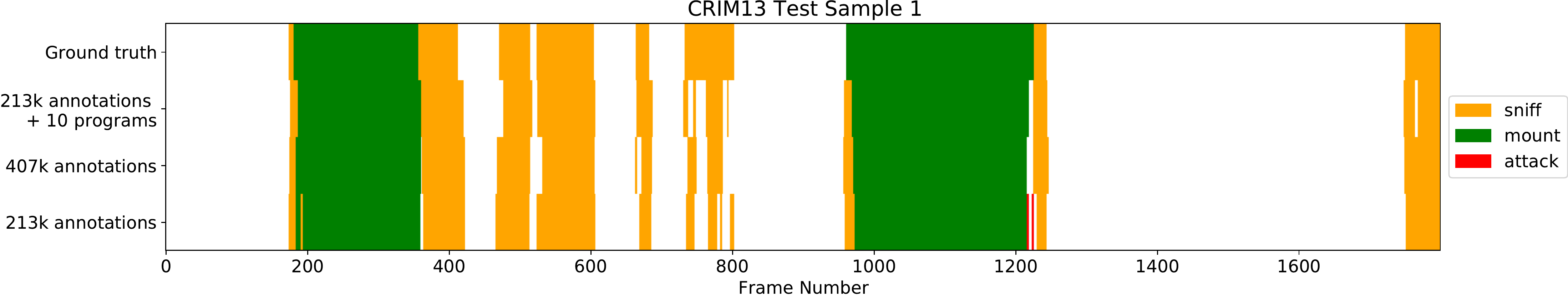}

\includegraphics[width=\textwidth]{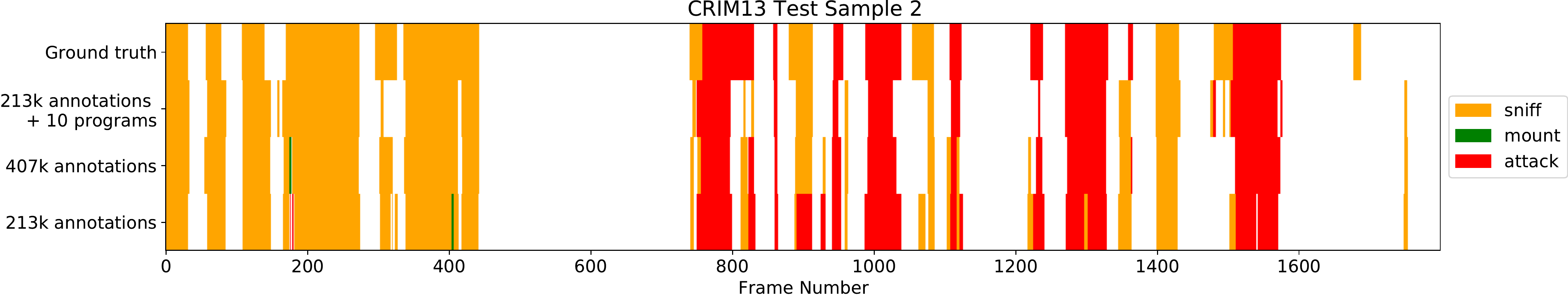}

\caption{ \textbf{Annotations on CRIM13 Test Set.} Classifier annotations on CRIM13 dataset vs. ground truth at 25Hz. For each sample, the second row corresponds to features + TREBA trained with 10 expert-engineered programs, with $\sim 50\%$ of the supervised behavior annotations. The third row corresponds to training using domain-specific features on the full dataset. The last row corresponds to training using domain-specific features on $\sim 50\%$ of data, without TREBA.}
\label{fig:sample_2}
\end{figure*}

\begin{figure*}

\centering
\includegraphics[width=\textwidth]{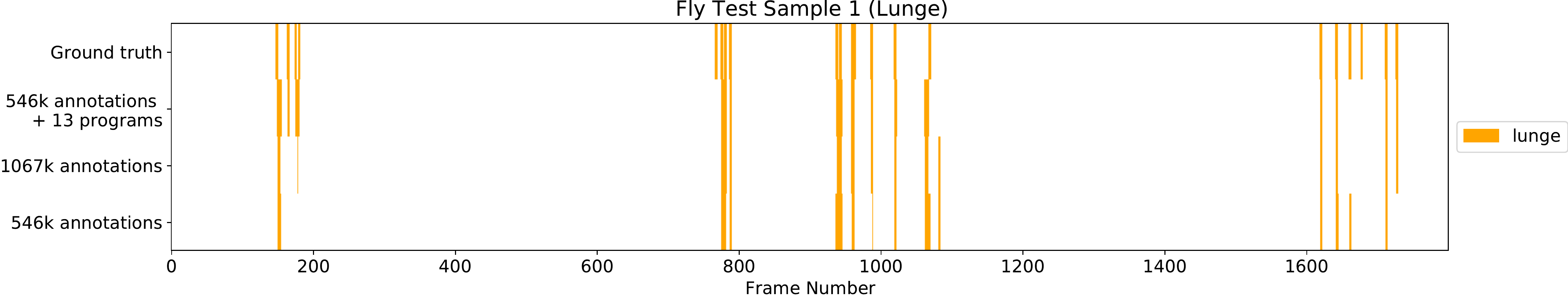}

\includegraphics[width=\textwidth]{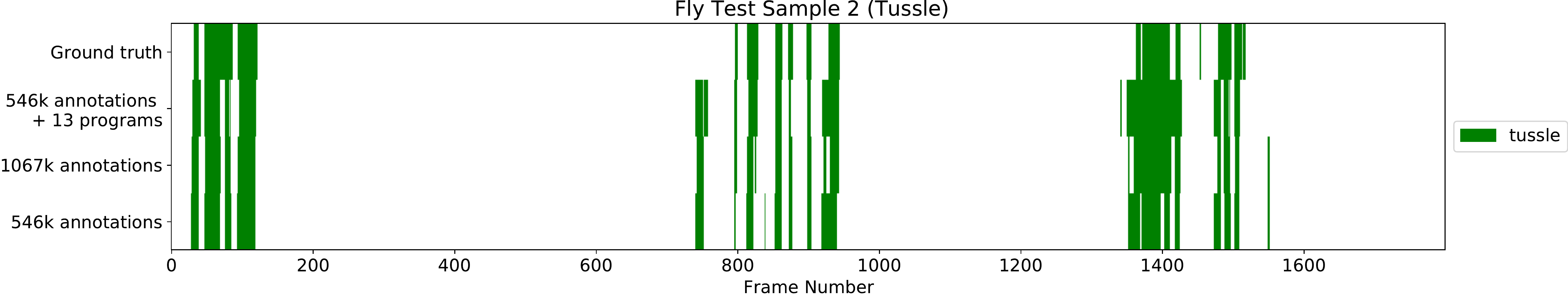}

\includegraphics[width=\textwidth]{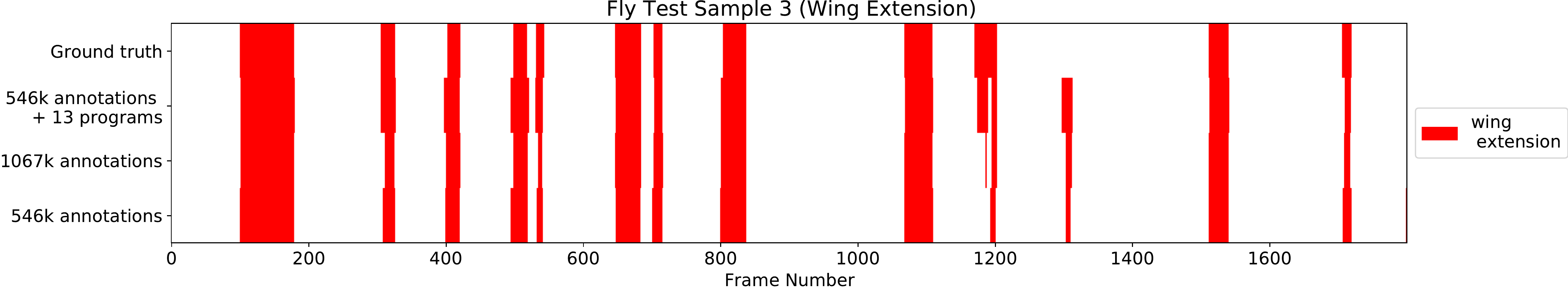}

\caption{ \textbf{Annotations on Fly Test Set.} Classifier annotations on Fly dataset vs. ground truth at 30Hz. For each sample, the second row corresponds to features + TREBA trained with 13 expert-engineered programs, with $\sim 50\%$ of the supervised behavior annotations. The third row corresponds to training using domain-specific features on the full dataset. The last row corresponds to training using domain-specific features on $\sim 50\%$ of data, without TREBA.}
\label{fig:sample_3}
\end{figure*}

We visualize classification samples for each dataset using input domain-specific features and TREBA (MARS in Figure~\ref{fig:sample_1}, CRIM13 in Figure~\ref{fig:sample_2}, Fly in Figure~\ref{fig:sample_3}). We visualize the classifier using TREBA at the training fraction such that the classifier MAP matches that of the fully-supervised baseline feature performance. Comparing the samples qualitatively, we see that the classifier output with the full training set is comparable to TREBA with $10\times$ reduced annotations on MARS and $2\times$ reduced annotations on CRIM13 and Fly. We note that the classifier trained on reduced data alone (last row of Figures~\ref{fig:sample_1}, ~\ref{fig:sample_2}, ~\ref{fig:sample_3}) is generally less accurate, compared to the classifiers trained with either full training data or with TREBA.

\end{document}